\documentclass[11pt]{article}

\usepackage[final]{acl}

\usepackage{times}
\usepackage{latexsym}

\usepackage[T1]{fontenc}

\usepackage[utf8]{inputenc}

\usepackage{microtype}

\usepackage{inconsolata}

\usepackage{graphicx}
\usepackage{amsmath}
\usepackage{tikz}
\usepackage{booktabs}
\usepackage{longtable}
\usetikzlibrary{positioning,arrows.meta,fit,backgrounds,shapes.geometric,calc}

\usepackage[colorinlistoftodos]{todonotes}

\title{METC: Multilingual Error Type Classifier for Grammatical Error Correction}

\author{
    Gabriele Matiddi\thanks{Work done while at Superhuman.}
    \quad Tetiana Drobot\textsuperscript{1}
    \quad Kostiantyn Omelianchuk\textsuperscript{1} \\
    \textsuperscript{1}Superhuman \\
    \texttt{gabrielematiddi@gmail.com} \\
    \texttt{\{tetiana.drobot,kostiantyn.omelianchuk\}@superhuman.com}
}

\hypersetup{
    pdftitle={METC: Multilingual Error Type Classifier
              for Grammatical Error Correction},
    pdfauthor={Gabriele Matiddi, Tetiana Drobot,
               Kostiantyn Omelianchuk}
}

\begin{document}
\maketitle
\begin{abstract}
Error-type annotation assigns a category to each grammatical error correction
edit. Existing ERRANT-style systems rely on language-specific linguistic tools
and rules, making extension to new languages difficult. We introduce and release \textbf{METC}, a multilingual error-type annotation system that combines a shared edit extractor, a multilingual token classifier, and a shared 42-label taxonomy without requiring language-specific linguistic pipelines. We introduce two 
complementary protocols to evaluate METC in six languages: (a) reference
matching measures how closely a system reproduces a corpus's chosen taxonomy;
and (b) LLM acceptability judging evaluates whether each system's
own label describes its edit, enabling direct comparison across incompatible
taxonomies. Direct acceptability judging places METC first among the published
systems in most evaluated languages and among the best in all of them. 
System-blind human studies in English and German validate the LLM evaluation,
with LLM--human label agreement comparable to inter-annotator agreement.
METC therefore provides a strong shared alternative to
language-specific error-type annotation pipelines.
\end{abstract}

\section{Introduction}
\label{sec:intro}

Error Type Annotation for Grammatical Error Correction assigns each
edit a label from a fixed taxonomy. The task has two independent dimensions:
\emph{edit extraction}, recovering which spans changed between a source and its
correction, and \emph{edit-type classification}, naming the error each edit
fixes. A correct span can carry a wrong label, and a wrong span a plausible
one, so the two are best assessed separately.
A widely used English system is ERRANT \citep{bryant-etal-2017-automatic}, a
rule-based classifier built over a language-specific linguistic pipeline
(\S\ref{sec:bg}). Adapting it to a new language requires substantial
language-specific work on that pipeline: published adaptations for German
and Russian do exactly that
\citep{boyd-2018-using,rozovskaya-roth-2019-grammar}, and a recent multilingual
ERRANT extension \citep{qiu-etal-2025-multilingual} retains per-language
components. Of the twelve learner corpora in MultiGEC-2025
\citep{masciolini-etal-2025-multigec}, the majority still have no published
Error Type Classifier.

\begin{figure*}[t]
\centering
\definecolor{errc}{rgb}{0.72,0.0,0.0}
\definecolor{okc}{rgb}{0.0,0.45,0.0}
\begin{tikzpicture}[
  font=\scriptsize,
  box/.style={draw, rounded corners=2pt, text width=26mm, align=left,
              inner sep=2mm, minimum height=18mm},
  oplbl/.style={font=\scriptsize\itshape, midway, above=0.8mm, align=center},
  arr/.style={-{Latex[length=2mm]}, semithick}
]
\node[box] (in)
  {{\footnotesize\bfseries Input}\\[3pt]
   \ttfamily I {\color{errc}dont} like\\ {\color{errc}this} cars .\\[2pt]
   \ttfamily I {\color{okc}don't} like\\ {\color{okc}these} cars .};
\node[box, right=8mm of in] (edit)
  {{\footnotesize\bfseries Extracted edits}\\[3pt]
   \ttfamily {\color{errc}dont}=>{\color{okc}don't}\\[2pt]
   \ttfamily {\color{errc}this}=>{\color{okc}these}};
\node[box, text width=40mm, right=8mm of edit] (dec)
  {{\footnotesize\bfseries Decorated input (one string)}\\[3pt]
   \ttfamily I [EDIT] {\color{errc}dont}=>{\color{okc}don't} [/EDIT] like
   [EDIT] {\color{errc}this}=>{\color{okc}these} [/EDIT] cars .};
\node[box, right=8mm of dec] (ann)
  {{\footnotesize\bfseries Labels (one pass)}\\[3pt]
   \ttfamily\bfseries Spelling/Misspelling\\[2pt]
   \ttfamily\bfseries Grammar/Agreement};
\draw[arr] (in) -- node[oplbl]{} (edit);
\draw[arr] (edit) -- node[oplbl]{} (dec);
\draw[arr] (dec) -- node[oplbl]{} (ann);
\end{tikzpicture}
\caption{The METC annotation pipeline on the worked example of
\S\ref{sec:metc-dmp}. Both extracted edits are decorated into a single input
string, and the token-classification model labels every marked span in one
pass.}
\label{fig:arch}
\end{figure*}
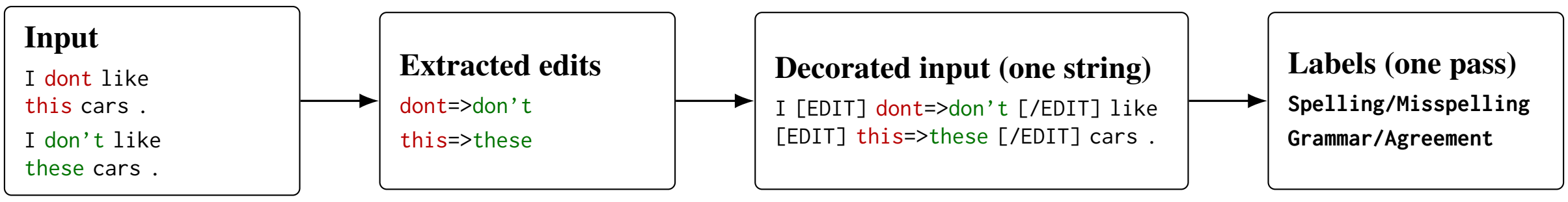

METC addresses the language-adaptation issue with a single multilingual
encoder over a shared, language-agnostic edits extractor
(\S\ref{sec:metc-dmp}). Figure~\ref{fig:arch} shows the complete pipeline:
METC extracts edits, decorates them in context, and labels every marked span in
one classifier pass. No language ID is passed at inference. All languages
share one classification head and the same fixed
script-aware edit-extraction heuristics.

\paragraph{Contributions.} Our contributions are twofold:

(i) \textbf{System}: we train and release METC under a
permissive license.\footnote{\url{https://github.com/superhuman/metc}}
Trained on over 1.5M examples spanning 200+ languages, METC combines a shared edit extractor, a multilingual labeler, and a shared 42-label taxonomy without language-specific linguistic pipelines or classification heads. We evaluate the resulting system across six languages.

(ii) \textbf{Evaluation}: we introduce and apply a two-part evaluation framework across six languages for comparing systems with incompatible taxonomies: (a) reference matching, which maps each
system's labels into a corpus taxonomy and measures agreement with the
reference annotation; and (b) mapping-free LLM acceptability evaluation, which
directly assesses whether each system's own label describes its edit. We
validate the LLM evaluation through system-blind human studies in English and
German, finding that LLM--human label agreement is comparable to
inter-annotator agreement.

\section{Background}
\label{sec:bg}

ERRANT combines a linguistically informed edit aligner
\citep{felice-etal-2016-automatic} with a rule-based classifier. The aligner
uses POS, lemma, and Damerau--Levenshtein features to form replacement edits;
the classifier applies rules over spaCy analyses\footnote{\url{https://spacy.io}}
and Hunspell lookups. Labels combine one of three operation prefixes,
\texttt{M} (missing), \texttt{U} (unnecessary), or \texttt{R} (replacement),
with one of 25 main classes and, where applicable, a subtype, yielding
approximately 55 labels on English data.

Adapting ERRANT requires a language-specific tokenizer, tagger, lemmatizer,
parser, morphological analyzer, spelling dictionary, and rule set. The
multilingual ERRANT of \citet{qiu-etal-2025-multilingual} replaces spaCy with
Stanza \citep{qi-etal-2020-stanza} but retains language-specific pipelines and
extension templates for its five reported languages.

Label inventories are fragmented as well. CoNLL-2014 defines 28 English types
\citep{ng-etal-2014-conll}, while other corpora use their own schemes, such as
MERLIN's \texttt{G\_Morphol\_Wrong/case} \citep{boyd-etal-2014-merlin} and
UA-GEC's \texttt{G/Case} \citep{syvokon-etal-2023-ua}.

Complementary work evaluates grammatical-error taxonomies themselves in terms
of exclusivity, coverage, balance, and usability
\citep{zou-etal-2025-revisiting}. Error-annotated learner corpora also include non-English resources such as COWS-L2H for Spanish L2 and heritage writing \citep{Yamada_Davidson_Fernández-Mira_Carando_Sagae_Sánchez-Gutiérrez_2020}.

\section{The METC system}
\label{sec:metc}

\subsection{Edits extraction}
\label{sec:metc-dmp}

The extractor turns a source and corrected sentence pair into edit spans in
four steps, all shared across languages.
\begin{enumerate}
\item \textbf{Tokenize.} Both strings are tokenized using the same fixed, script-aware procedure. Chinese characters and Japanese kana become individual tokens. Other text, including Hangul, Thai, and alphabetic scripts, remains grouped until whitespace, selected punctuation, or emoji characters. The extractor preserves casing and does not apply general Unicode normalization, although selected whitespace characters are normalized. Mixed-script text may remain within a single token, and multi-code-point emoji sequences may be split. This procedure avoids the need for language-specific linguistic tokenizers.
\item \textbf{Diff.} Diff Match Patch (DMP; \citealp{dmp}) runs over the two token sequences
and returns runs of kept, deleted, and inserted tokens.
\item \textbf{Merge.} Neighboring deleted and inserted runs become one
replacement edit, so a correction reads as one change rather than a deletion
plus an insertion.
\item \textbf{Split edge punctuation.} When one replacement changes edge
punctuation together with the word next to it, the punctuation becomes its own
edit. This is the only change made to edit spans after the diff, and it exists
because the two changes are different error types and deserve separate labels.
\end{enumerate}
A short example carries through the steps. For the pair
\textit{``I dont like this cars .''} $\to$ \textit{``I don't like these cars
.''}, the diff marks \texttt{dont} and \texttt{this} as deleted and
\texttt{don't} and \texttt{these} as inserted, and the merge turns them into
the two replacements \texttt{\{dont=>don't\}} and \texttt{\{this=>these\}};
no edge punctuation is involved, so the last step changes nothing. That step
fires in pairs like \texttt{' today} $\to$ \texttt{" Today}. Before splitting,
the merged replacement is \texttt{\{' today=>" Today\}}, which combines a
quotation-mark change with a casing change. It splits into
\texttt{\{'=>"\}} and \texttt{\{today=>Today\}}, later labeled as one
Punctuation and one Spelling error rather than a single compound edit.

\subsection{Edits labeling}
\label{sec:metc-labeling}

Labeling runs in two steps. First, decoration: the source text is rewritten
with every extracted span wrapped in marker tokens, replacement included and
surrounding context intact. The example above becomes
\texttt{I [EDIT] dont=>don't [/EDIT] like [EDIT] this=>these [/EDIT] cars .}
Second, classification: the token-classification model runs once over the
decorated string, and one of the 42 labels is read at each opening
\texttt{[EDIT]} token. Because every edit contributes exactly one opening
marker, its prediction maps one-to-one to the complete enclosed edit,
regardless of how many tokens the edit contains. For the
example, the model returns \texttt{Spelling/Misspelling} for the first span and
\texttt{Grammar/Agreement} for the second;
\hyperref[fig:arch]{Figure~\ref{fig:arch}} traces this example end to end.

\subsection{Taxonomy design}
\label{sec:metc-taxonomy}

METC's 42-label taxonomy is conceptually hierarchical, de facto flat at
inference. Seven descriptive semantic groups organize its labels: Grammar,
Formatting, Spelling, Punctuation, Rephrasing, SentenceStructure, and Other.
Inference predicts one label per edit, with no hierarchical decision,
per-group calibration, or language-specific head. The full inventory is:

\begin{itemize}
\setlength\itemsep{2pt}
\item \textbf{Formatting (9):} AbbreviationFormat, CurrencyFormat, DateFormat,
  MeasurementFormat, MissingSpace, Other, RedundantSpace, SpellOutNumber, TimeFormat.
\item \textbf{Grammar (14):} Adjective, Adverb, Agreement, Conjunction, Determiner,
  Morphology, Negation, Noun, Particle, Preposition, Pronoun, Verb, VerbSVA,
  VerbTenseForm.
\item \textbf{Other (3):} Compound\_Trivial, Compound\_Moderate, Compound\_Complex
  (multi-issue edits at three increasing levels of severity; see Table \ref{tab:metc-desc}).
\item \textbf{Punctuation (3):} ClosingPunctuation, Other, SentenceSplit.
\item \textbf{Rephrasing (4):} Formality, Other, Tone, Wordiness.
\item \textbf{SentenceStructure (2):} SentenceFragment, WordOrder.
\item \textbf{Spelling (7):} Diacritic, Merge, Misspelling, Other,
  SpellConfused\_Capitalize, SpellConfused\_Confused, SpellConfused\_Lowercase.
\end{itemize}

Labels are represented as \texttt{group/sublabel} (for example,
\texttt{Grammar/VerbTenseForm}); numerical IDs correspond to the released
label mapping.

The taxonomy builds on ERRANT's label inventory, whose structure
\S\ref{sec:bg} describes. We retain its
part-of-speech-anchored categories, including particle, word order,
morphology, and subject-verb agreement.

We depart from ERRANT in five ways. (i) We collapse
the operation dimension into a single label per category: the operation is
recoverable from the edit itself. (ii) We reorganize the hierarchy, merging
ERRANT's fine morphological subtypes (verb form, inflection, and tense become
one verb-tense-form category) while adding distinctions where ERRANT is
coarse. (iii) We refine categories that ERRANT bundles: punctuation splits
into sentence-final and sentence-boundary types; spelling into misspelling,
confusion, and diacritic types; and ERRANT's orthography category, which
mixes casing and whitespace, is separated by function into spelling and
formatting, with formatting promoted to a top-level label covering dates,
currency, measurement units, and number spell-out. (iv) Because the taxonomy
is multilingual rather than English-only, categories are assigned by
grammatical function rather than English part of speech; the particle
category is generalized from English phrasal-verb particles to
cross-linguistic particles, and a diacritic category with no ERRANT
counterpart is added. Both changes were finalized after a manual
taxonomy-development review of Slovak, Polish, Korean, Ukrainian, and Arabic
data revealed systematic edits that the English-derived scheme could not
express. (v) Where ERRANT
sends meaning-preserving edits to a catch-all other category, we separate them
into a dedicated Rephrasing branch (wordiness, formality, and tone) and add
a sentence-fragment category and a graded Compound label for edits that span
multiple error types. Appendix~\ref{app:errant-metc} gives the
category-by-category correspondence.

\section{Training approach}
\label{sec:training}

METC is trained as a student model over a teacher-annotated corpus. A
GPT-5.4\footnote{Model ID \texttt{gpt-5.4}:
\url{https://developers.openai.com/api/docs/models/gpt-5.4}}
annotation teacher applies a multi-edit prompt to spans derived from existing
source--correction pairs. The student uses the
\texttt{jhu-clsp/mmBERT-small} encoder \citep{marone-etal-2025-mmbert}, selected
for broad multilingual coverage and its small encoder configuration; we did
not compare alternative encoders in this study. It is fine-tuned with
cross-entropy over the 42-label taxonomy.

\subsection{Training data annotation}
\label{sec:training-annot}

The proprietary training corpus holds approximately 1.7M source--target pairs
spanning 200+ languages. In each pair, the source is a sentence containing one or more errors, and the target is its corrected version. The corpus is
divided into English, five
higher-resource languages (French, Spanish, Italian, Portuguese, and German),
and a long-tail multilingual set in which each language is capped at 25K
examples. METC's language-agnostic edits extractor converts each pair into
edit spans, and a separate GPT-5.4 annotation teacher assigns each edit one of
the 42 shared labels under one cross-lingual prompt, which we release with the
model checkpoint and code. We do not release our proprietary training corpus; however, we believe that a comparable corpus can be generated by applying the METC-Teacher prompt with LLMs to any publicly available grammatical error correction dataset;
Table~\ref{tab:training-dist} reports the training mixture's top-level
error-type distribution. The long-tail tier is the largest component of the
corpus, and its label distribution differs from English: spelling is most
common in the long-tail data, whereas grammar dominates English.

\begin{table}[t]
\centering
\footnotesize
\setlength{\tabcolsep}{3pt}
\begin{tabular}{l r r r r}
\toprule
\textbf{Category} & \textbf{Overall} & \textbf{Eng.} & \textbf{5\,langs} & \textbf{Long-tail} \\
\midrule
Spelling           & 30.4 & 10.0 & 32.5 & 37.5 \\
Punctuation        & 24.0 & 24.1 & 23.6 & 24.2 \\
Grammar            & 22.5 & 40.5 & 23.8 & 15.3 \\
Formatting         & 10.8 &  6.2 & 12.1 & 12.2 \\
Other              &  8.5 &  7.2 &  6.6 &  9.7 \\
Rephrasing         &  2.8 & 10.6 &  0.4 &  0.6 \\
SentenceStructure  &  0.9 &  1.5 &  1.0 &  0.7 \\
\midrule
Edits/sentence     & 2.69 & 2.28 & 2.07 & 3.25 \\
\bottomrule
\end{tabular}
\caption{Top-level training-corpus distribution (\% of edits) by language
tier. Tier sizes are 438K English rows, 450K rows across the five
higher-resource languages, and 811K rows across 200+ long-tail languages. The
last row gives mean edits per sentence.}
\label{tab:training-dist}
\end{table}

\subsection{Student architecture and fine-tuning}
\label{sec:training-recipe}

Each training example uses the same decorated sentence representation as at
inference. The
teacher label for an edit is assigned to its opening \texttt{[EDIT]} token;
all other token positions are ignored by the loss. One forward pass therefore
classifies every marked edit in the sentence.

We combine all examples from the English, higher-resource, and long-tail
groups described above. After shuffling the combined dataset with seed 42, we
use 90\% for training and hold out 10\% for validation. This yields
1,529,258 training examples and 169,918 validation examples. Checkpoints are
compared only by
validation macro-F1 on that internal split; none of the seven paper benchmarks
is used for checkpoint selection. The selected checkpoint is step 23,000,
with validation macro-F1 0.7583. Training stopped at step 26,000 after three
evaluations without improvement. Appendix~\ref{app:hyperparams} gives the
remaining hyperparameters. 
Versions of METC have also been used as part of Grammarly's grammatical error correction pipeline across 24 languages. Appendix~\ref{app:language_coverage} provides further details on training-language coverage and production use.

\section{Evaluation approaches}
\label{sec:evaluation}
\subsection{Background and motivation}
\label{sec:eval-background}
There is no canonical error-type taxonomy: each corpus defines its own scheme
(\S\ref{sec:bg}). Reference-matching therefore measures agreement with
a chosen corpus after mapping between taxonomies, not absolute correctness; a
valid label can be penalized when that mapping collapses a distinction. We read
a higher score as a smaller taxonomy shift, not proof of a better classifier.

We therefore use two complementary protocols: reference-matching quantifies
compatibility with a corpus taxonomy, while an acceptability judge assesses
each system's own label claim without cross-taxonomy mapping.
Table~\ref{tab:evaluation-protocols} contrasts their questions, outputs, and
main risks.
\begin{table*}[t]
\centering
\small
\begin{tabular}{p{0.17\linewidth}p{0.36\linewidth}p{0.36\linewidth}}
\toprule
 & \textbf{Reference-matching} & \textbf{Acceptability judging} \\
\midrule
Question &
Does the system reproduce the corpus reference annotation? &
Is the system's own claim about the edit adequate? \\
Reference &
Corpus-native reference annotation &
Acceptability judge \\
Adapter &
Map the system-native label into the corpus taxonomy &
Render the system-native label as a neutral English description \\
Unit &
Sentence-level aggregation of TP, FP, and FN &
Edit-level Good, Acceptable, or Bad judgment \\
Output &
Detection and classification $F_{0.5}$ &
Span quality and label quality percentages \\
Main risk &
Target-taxonomy choice can collapse native distinctions &
Judge quality and description-rendering choices can affect verdicts \\
\bottomrule
\end{tabular}
\caption{The two evaluation protocols. Reference-matching reports a
corpus-relative reference score. Acceptability judging reduces the
target-taxonomy effect by evaluating each system's own label claim directly.}
\label{tab:evaluation-protocols}
\end{table*}
\subsection{Datasets and systems}
\label{sec:eval-data}
We evaluate on seven dataset settings spanning six languages: English
CoNLL-2014 \citep{ng-etal-2014-conll} (annotators~0 and~1), German MERLIN
\citep{boyd-etal-2014-merlin}, Czech MERLIN
\citep{boyd-etal-2014-merlin}, Korean KoLLA v2
\citep{park-2025-kolla-v2}, Ukrainian UA-GEC
\citep{syvokon-etal-2023-ua}, and Chinese MuCGEC
\citep{zhang-etal-2022-mucgec}. Each corpus's annotations define the reference
label space for its setting. MuCGEC has no error-type labels, so we derive
them with ChERRANT \citep{zhang-etal-2022-mucgec}; KoLLA v2 similarly uses
KAGAS-derived labels \citep{yoon-etal-2023-towards}.

Both protocols compare METC with the systems available for each dataset:
ERRANT and jp\_errant for English; the German ERRANT modification (gERRANT)
and jp\_errant \citep{qiu-etal-2025-multilingual} for German; and jp\_errant
for Czech, Korean, Ukrainian, and Chinese. We also include METC-teacher, the
teacher configuration used to produce the training annotations, applied to
METC's edits. Student and teacher label identical spans, so their detection
scores coincide and their classification difference isolates the effect of
distillation. Appendix~\ref{app:boyd-port} documents the gERRANT rebuild,
Appendix~\ref{app:mappings} gives the classification mappings, and
Appendix~\ref{app:artifacts} reports dataset statistics and licenses.
\subsection{Reference-matching}
\label{sec:eval-reference}
\label{sec:results-reference}
The reference-matching protocol reports two $F_{0.5}$ scores, following the
edit-level precision and recall formulation of MaxMatch ($M^2$;
\citealp{dahlmeier-ng-2012-better}) and ERRANT
\citep{bryant-etal-2017-automatic}. Every system annotates the same fixed
source--correction pair, and we score how closely its edit spans and labels
match the corpus annotation. Standard $M^2$ may choose the segmentation that
best matches the reference. We keep each system's segmentation fixed because
span choice is part of the comparison. \emph{Detection} finds the largest
one-to-one matching between system and reference edits whose source spans
overlap. Unmatched system and reference edits count as false positives and
false negatives, respectively.
\emph{Classification} uses the same matching procedure but additionally
requires the predicted error-type label to match the reference label. METC,
ERRANT, the German ERRANT modification, jp\_errant, and the corpus annotations use
different taxonomies, so each system's
native labels are mapped into the corresponding corpus-native taxonomy
before classification.
These mappings are evaluation artifacts, not part of METC. All reported
classification results use the same fixed source-to-target mappings
(~\ref{app:mappings}). The mapping step introduces a target-taxonomy
effect: a semantically valid native label can
still be penalized when the mapping merges or redirects it.
This dependency is the protocol's main limitation: mappings must be
hand-built for every system and corpus pair, and a differing native taxonomy
is penalized for the difference itself; acceptability judging is included to
complement this.
Confidence intervals are computed by paired sentence-level
bootstrap resampling with 1,000 samples and seed 42.
Appendix~\ref{app:exact-span} repeats the comparison with exact span equality.

Table~\ref{tab:exp-h} separates two sources of reference disagreement:
panel~(a) compares edit-span detection, while panel~(b) additionally requires
the mapped label to match. On English, both baselines are closer to CoNLL than
METC. The ERRANT--METC gap widens when label agreement is required, indicating
a taxonomy effect; the jp\_errant--METC gap narrows. METC is closer than the
German ERRANT modification on German and jp\_errant on Czech, whereas
jp\_errant is closer on Ukrainian. jp\_errant's near-perfect Korean
classification score reflects shared KAGAS annotations rather than an
independent comparison. On Chinese, METC is closer to the ChERRANT-derived
labels even though jp\_errant shares that label space.
\begin{table*}[t]
\centering
\small
\setlength{\tabcolsep}{2.5pt}
{\bfseries (a) Detection\par}
\vspace{0.15em}
\begin{tabular*}{\textwidth}{@{\extracolsep{\fill}}lrrrrrrr@{}}
\toprule
System & EN0 & EN1 & DE & CS & KO & UK & ZH \\
\midrule
METC       & 0.871\,$\pm$\,0.014 & 0.850\,$\pm$\,0.014 & \textbf{0.648\,$\pm$\,0.037} & \textbf{0.681\,$\pm$\,0.012} & 0.970\,$\pm$\,0.009 & 0.760\,$\pm$\,0.014 & \textbf{0.871\,$\pm$\,0.009} \\
ERRANT     & \textbf{0.917\,$\pm$\,0.011} & \textbf{0.924\,$\pm$\,0.011} & -- & -- & -- & -- & -- \\
gERRANT    & -- & -- & 0.608\,$\pm$\,0.034 & -- & -- & -- & -- \\
jp\_errant & \textbf{0.916\,$\pm$\,0.011} & 0.920\,$\pm$\,0.011 & 0.609\,$\pm$\,0.034 & 0.655\,$\pm$\,0.013 & \textbf{0.999\,$\pm$\,0.001} & \textbf{0.916\,$\pm$\,0.009} & \textbf{0.871\,$\pm$\,0.009} \\
\bottomrule
\end{tabular*}

\vspace{0.55em}
{\bfseries (b) Classification\par}
\vspace{0.15em}
\begin{tabular*}{\textwidth}{@{\extracolsep{\fill}}lrrrrrrr@{}}
\toprule
System & EN0 & EN1 & DE & CS & KO & UK & ZH \\
\midrule
METC       & 0.527\,$\pm$\,0.022 & 0.556\,$\pm$\,0.018 & \textbf{0.368\,$\pm$\,0.026} & \textbf{0.442\,$\pm$\,0.011} & 0.362\,$\pm$\,0.023 & 0.438\,$\pm$\,0.019 & 0.538\,$\pm$\,0.016 \\
METC-teacher & 0.493\,$\pm$\,0.022 & 0.527\,$\pm$\,0.018 & \textbf{0.371\,$\pm$\,0.026} & \textbf{0.446\,$\pm$\,0.011} & 0.356\,$\pm$\,0.022 & 0.491\,$\pm$\,0.019 & \textbf{0.585\,$\pm$\,0.017} \\
ERRANT     & \textbf{0.595\,$\pm$\,0.021} & \textbf{0.654\,$\pm$\,0.021} & -- & -- & -- & -- & -- \\
gERRANT    & -- & -- & 0.311\,$\pm$\,0.023 & -- & -- & -- & -- \\
jp\_errant & 0.550\,$\pm$\,0.022 & 0.605\,$\pm$\,0.018 & 0.178\,$\pm$\,0.016 & 0.152\,$\pm$\,0.007 & \textbf{0.999\,$\pm$\,0.001} & \textbf{0.545\,$\pm$\,0.015} & 0.459\,$\pm$\,0.016 \\
\bottomrule
\end{tabular*}
\caption{Reference-matching $F_{0.5}$ by system and dataset. Panel~(a) measures
detection by matching spans without labels; panel~(b) measures classification
by additionally requiring the mapped label to match the corpus-native taxonomy.
Values show 95\% paired sentence-bootstrap intervals as $\pm$ half-width. Within
each panel and dataset, bold marks the highest score and systems whose paired
difference from it includes zero; ``--'' marks an unavailable system. EN0 and
EN1 are CoNLL-2014 annotators 0 and 1. METC-teacher is omitted from panel~(a)
because it shares METC's extractor. Korean jp\_errant in panel~(b) shares the
KAGAS annotation scheme with KoLLA and is not an independent classification
comparison.}
\label{tab:exp-h}
\end{table*}
\subsection{Acceptability judging}
\label{sec:eval-judge}
\label{sec:results-judge}
Acceptability judging uses the established LLM-as-judge paradigm
\citep{liu-etal-2023-g,NEURIPS2023_91f18a12} to score each edit directly,
without the cross-taxonomy mapping. For each edit, an LLM judge sees the source
sentence, the corrected sentence, the edit as
source-to-target text, and a one-sentence English description of the system's
label~(Appendix~\ref{app:descriptions}); it does not see the system name, the native label code, or the taxonomy
prefix. It returns two ratings, each Good, Acceptable, or Bad with a short
rationale (following \citealp{bryant-etal-2017-automatic}): \emph{span quality},
whether the edit's span pinpoints the error, and \emph{label quality}, whether
the description accurately characterizes the edit. The runs reported below
use GPT-5.4, the original evaluation judge, with strict JSON output; the
percentages are point estimates. We then repeated the full run with
Gemini~3.5 Flash\footnote{Model ID \texttt{gemini-3.5-flash}:
\url{https://ai.google.dev/gemini-api/docs/models/gemini-3.5-flash}},
a model from another family, as a cross-family sensitivity check
(\S\ref{sec:results-judge-robustness}). GPT-5.4 remains the primary judge
because it was used in the initial analysis; this designation is not evidence
that either judge better matches humans.

Table~\ref{tab:llm-summary} summarizes the primary-judge results. The edit
counts and full outcome distributions appear in
Tables~\ref{tab:exp-l} and~\ref{tab:exp-sent} in the appendix. METC usually
produces fewer, larger edits than the comparison systems, while its span-Good
rates remain broadly comparable. Under the primary judge, METC has a higher label-Good rate than
jp\_errant in every setting except Ukrainian; the German ERRANT modification
is higher on German. On English, this ordering contrasts with
reference-matching, where the baselines are closer to CoNLL, illustrating that
the protocols measure different properties.
\begin{table*}[t]
\centering
\small
\setlength{\tabcolsep}{2.5pt}
{\bfseries (a) Edit-level Good: span / label (\%)\par}
\vspace{0.15em}
\begin{tabular*}{\textwidth}{@{\extracolsep{\fill}}lrrrrrrr@{}}
\toprule
System & EN0 & EN1 & DE & CS & KO & UK & ZH \\
\midrule
Reference      & 83.4 / 69.7 & 87.1 / 78.8 & 76.4 / 59.3 & 85.2 / 60.5 & 77.5 / 39.8 & 79.4 / 74.3 & 69.1 / 39.9 \\
METC           & 74.8 / \textbf{63.7} & 77.1 / \textbf{66.7} & 79.4 / 72.4 & 83.9 / 62.5 & 70.2 / \textbf{57.3} & 82.1 / 64.7 & 67.4 / 48.8 \\
METC-teacher   & 74.8 / \textbf{64.0} & 77.1 / \textbf{66.7} & 78.8 / 72.1 & 84.4 / \textbf{66.4} & 72.5 / \textbf{58.4} & 81.2 / \textbf{78.7} & 67.1 / \textbf{50.3} \\
ERRANT/gERRANT & 78.6 / 60.7 & 78.9 / 62.5 & 84.8 / \textbf{75.0} & -- & -- & -- & -- \\
jp\_errant     & 79.0 / 59.0 & 79.0 / 60.9 & 81.8 / 47.8 & 87.1 / 26.0 & 77.6 / 40.1 & 71.0 / 64.9 & 63.1 / 31.8 \\
\bottomrule
\end{tabular*}

\vspace{0.55em}
{\bfseries (b) Sentence-level adequacy (\%)\par}
\vspace{0.15em}
\begin{tabular*}{\textwidth}{@{\extracolsep{\fill}}lrrrrrrr@{}}
\toprule
System & EN0 & EN1 & DE & CS & KO & UK & ZH \\
\midrule
Reference      & 61.2$\pm$2.5 & 63.4$\pm$2.4 & 47.4$\pm$3.5 & 44.2$\pm$1.4 & 35.2$\pm$2.6 & 57.0$\pm$2.1 & 26.4$\pm$1.9 \\
METC           & \textbf{59.3$\pm$2.6} & \textbf{52.9$\pm$2.5} & 60.9$\pm$2.8 & 52.3$\pm$1.2 & 50.0$\pm$2.8 & 50.8$\pm$2.1 & 29.3$\pm$2.3 \\
METC-teacher   & \textbf{58.5$\pm$2.6} & \textbf{52.9$\pm$2.5} & 61.2$\pm$2.8 & \textbf{56.9$\pm$1.1} & \textbf{54.3$\pm$2.7} & \textbf{59.2$\pm$2.1} & \textbf{32.0$\pm$2.3} \\
ERRANT/gERRANT & 55.7$\pm$2.7 & 50.1$\pm$2.6 & \textbf{65.5$\pm$2.7} & -- & -- & -- & -- \\
jp\_errant     & 55.0$\pm$2.6 & 47.8$\pm$2.4 & 39.9$\pm$2.6 & 27.7$\pm$0.9 & 35.5$\pm$2.6 & 47.3$\pm$2.2 & 15.4$\pm$1.8 \\
\bottomrule
\end{tabular*}
\caption{Primary-judge summary by system and dataset. Panel~(a) reports the
shares of edits rated Good on span and label quality, shown as span / label;
these values are point estimates. Panel~(b) reports sentence adequacy,
$\mathrm{Perfect}+0.5\cdot\mathrm{Acceptable}$, with 95\%
sentence-bootstrap intervals shown as $\pm$ half-width. Bold applies only to
label quality in panel~(a) and to adequacy in panel~(b); it marks the highest
score among compared systems together with systems whose paired difference
from it includes zero. Reference rows are not bolded. ``--'' marks an
unavailable system. Reference denotes the corpus annotation, including KAGAS
for Korean and ChERRANT for Chinese. Full distributions and counts appear in
Tables~\ref{tab:exp-l} and~\ref{tab:exp-sent} in the appendix.}
\label{tab:llm-summary}
\end{table*}
\subsection{Sentence-level adequacy}
\label{sec:results-sentence}
The acceptability judge (\S\ref{sec:results-judge}) rates each edit on two
separate axes, and systems differ in how many edits they propose per sentence,
which makes the cumulative effect of a system's span and label decisions hard
to read off the per-axis percentages. We therefore aggregate the per-edit
judgments into one sentence-level score. A sentence is Perfect if every
proposed edit is Good on both axes, Acceptable if no edit is Bad on either
axis, and Bad otherwise. Adequacy is
$\mathrm{Perfect}+0.5\cdot\mathrm{Acceptable}$. We use sentence-bootstrap
intervals matching the resampling used for the reference-matching scores.
Because the judge only rates
edits a system actually proposes, the metric ranges over the sentences in
which a system makes at least one edit and does not account for unproposed
edits.

Panel~(b) of Table~\ref{tab:llm-summary} reports the result. Under the primary judge, METC has
the highest adequacy score among student systems on every setting except
German, where gERRANT is higher. METC-teacher has the highest score on Czech,
Korean, Ukrainian, and Chinese. On 909 shared German sentences, gERRANT has
4.51 points higher adequacy than METC under a paired sentence-level bootstrap
(95\% interval [2.09, 6.88]; 1{,}000 resamples, seed 42). The paired interval
excludes zero even though the marginal intervals in
panel~(b) overlap.

\subsection{Cross-judge robustness}
\label{sec:results-judge-robustness}

To check whether the results depend on the primary judge, we repeated the
complete evaluation with Gemini~3.5 Flash. The judges agree on
74.0--90.1\% of span ratings and 60.2--76.6\% of label ratings (see Table \ref{tab:second-judge}). The broad
picture remains the same: METC retains clear label-quality leads over
jp\_errant on Czech and Korean, the teacher remains ahead of the student on
Ukrainian, and gERRANT remains ahead on German sentence adequacy. Some closer
comparisons change, particularly on Ukrainian. Neither judge matches the human
ratings better across both languages and rating axes. We therefore base our
main conclusions on results that hold across judges and use human evaluation
to interpret close comparisons.

\begin{table*}[!t]
\centering
\small
\begin{tabular}{lrrrrrrr}
\toprule
 & \multicolumn{2}{c}{Agreement \%} & \multicolumn{5}{c}{Label-Good \% (primary / second judge)} \\
\cmidrule(lr){2-3} \cmidrule(lr){4-8}
Setting & Span & Label & Reference & METC & METC-teacher & ERRANT/gER & jp\_errant \\
\midrule
EN ann0 & 84.5 & 73.7 & 69.7 / 76.8 & 63.7 / 72.5 & 64.0 / 71.2 & 60.7 / 69.0 & 59.0 / 68.8 \\
EN ann1 & 84.8 & 75.9 & 78.8 / 85.5 & 66.7 / 79.2 & 66.7 / 77.7 & 62.5 / 73.9 & 60.9 / 73.1 \\
DE      & 86.1 & 76.6 & 59.3 / 67.3 & 72.4 / 79.2 & 72.1 / 81.3 & 75.0 / 80.6 & 47.8 / 70.8 \\
CS      & 90.1 & 70.3 & 60.5 / 72.8 & 62.5 / 78.1 & 66.4 / 82.5 & --          & 26.0 / 54.4 \\
KO      & 81.4 & 66.3 & 39.8 / 51.9 & 57.3 / 66.2 & 58.4 / 72.2 & --          & 40.1 / 51.9 \\
UK      & 81.0 & 75.3 & 74.3 / 79.0 & 64.7 / 64.2 & 78.7 / 79.5 & --          & 64.9 / 78.6 \\
ZH      & 74.0 & 60.2 & 39.9 / 67.4 & 48.8 / 66.7 & 50.3 / 72.4 & --          & 31.8 / 52.3 \\
\bottomrule
\end{tabular}
\caption{Cross-judge robustness. Agreement is the share of edits receiving the
same rating from both judges. Label-Good rates are shown as primary / second
judge.}
\label{tab:second-judge}
\end{table*}

\paragraph{Where judges differ.}
Most differences occur for transition-style descriptions used by
jp\_errant's multilingual variants. For \textit{Man} $\to$ \textit{Mann}, for
example, the label describes ``a pronoun corrected to a noun.'' The primary
judge treats the underlying error as spelling, while the second accepts the
description as a literal account of the surface change. This example shows why
transition-style labels are more sensitive to the choice of judge.

\subsection{Human evaluation}
\label{sec:results-human}

Twelve professional annotators applied the span- and label-quality protocol to
a system-blind English sample of 800 items (200 per system); each item was
assigned independently to two annotators. Of 1,600 assignments, 1,290 were
fully scorable, 309 were marked unannotatable, and one was missing from the
export. For German, we annotated the first 200 items from a deterministically
shuffled pool containing 200 items per system. The realized sample contains 54
reference, 48 METC, 48 gERRANT, and 50 jp\_errant items.
Each item received two independent assignments; 389 of the 400 assignments
were fully scorable and 11 were marked unannotatable.
Table~\ref{tab:human-eval} reports edit-level rates and sentence adequacy for
both studies.

\begin{table*}[!t]
\centering
\small
\begin{tabular}{lrrrrr}
\toprule
 & Span \% & \multicolumn{3}{c}{Label \%} & \\
\cmidrule(lr){3-5}
System & Good & Good & Accept. & Bad & Adequacy \\
\midrule
\multicolumn{6}{l}{\textit{English}} \\
Reference & 93.7 & 70.9 & 19.7 & 9.4 & 64.7\,$\pm$\,5.3 \\
METC & 90.6 & 67.4 & 24.7 & 7.8 & 65.7\,$\pm$\,4.7 \\
ERRANT & 93.9 & 71.1 & 21.4 & 7.5 & 66.1\,$\pm$\,5.0 \\
jp\_errant & 91.3 & 70.3 & 22.7 & 7.0 & 63.5\,$\pm$\,4.9 \\
\midrule
\multicolumn{6}{l}{\textit{German}} \\
Reference & 100.0 & 82.5 & 11.2 & 6.3 & 82.9\,$\pm$\,6.0 \\
METC & 96.8 & 78.7 & 14.2 & 7.1 & 69.3\,$\pm$\,7.0 \\
gERRANT & 97.0 & 86.5 & 8.6 & 4.9 & 73.4\,$\pm$\,8.6 \\
jp\_errant & 97.4 & 57.3 & 29.6 & 13.1 & 50.5\,$\pm$\,9.0 \\
\bottomrule
\end{tabular}
\caption{Human evaluation on English and German. Span and label rates pool
3,505 English and 1,029 German scorable edit verdicts. Adequacy first averages
fully scorable assignment scores within each item and then averages items,
with 95\% item-bootstrap intervals shown as $\pm$ half-width. Adequacy item
counts in table order are $n=180,174,182,189$ for English and
$n=54,48,48,50$ for German.}
\label{tab:human-eval}
\end{table*}

On the 1,901 unique scorable English edits, the two LLM judges agree exactly on
84.6\% of span verdicts and 74.5\% of label-quality verdicts. On the 535 German
edits, agreement is 89.0\% and 78.1\%, respectively. Across the two studies,
LLM--human label agreement ranges from 62.3\% to 70.9\%, comparable to
human--human agreement of 65.1\% on English and 64.2\% on German. These
agreement levels support LLM judging as an efficient method for broad
comparisons, with human annotation calibrating system-level conclusions.

On English, professional annotation places METC slightly below ERRANT while
both LLM judges place it slightly above. Because the human label-quality and
adequacy intervals overlap, we report the two systems as comparable. On
German, human annotators rate gERRANT's labels higher (86.5\% vs. 78.7\%). Its
sentence-adequacy point estimate is also higher (73.4\% vs. 69.3\%), but that
difference is not statistically significant. The larger long-tail and
Ukrainian teacher--student differences remain the more reliable observations.

\section{Conclusion}
\label{sec:conclusion}
We introduced METC, a shared multilingual error-type annotation system combining an edit extractor, a multilingual token classifier, and a shared 42-label taxonomy. Across seven benchmark settings in six languages, the results support sharing these components without rebuilding a linguistic pipeline for each language. The two evaluation
protocols answer complementary questions: reference-matching measures
compatibility with a selected benchmark taxonomy, while acceptability judging
tests whether each system's own labels describe its edits.

The results also delimit that claim. On English, ERRANT is closer to CoNLL
under reference-matching, but the independently repeated human evaluation
produces overlapping label-quality estimates for METC and the English
comparison systems. On German, human annotators give the dedicated ERRANT
modification higher label-quality and sentence-adequacy point estimates, but
the sentence-level difference is not statistically significant. On
Ukrainian, the gap
between METC-teacher and the student identifies distillation as a remaining
issue. The Chinese comparison is less direct because its reference labels are
generated automatically with ChERRANT.

\paragraph{Future work.} Useful directions for further improving METC include
refreshing teacher-labeled data where the student lags, particularly for
Ukrainian, and testing or extending the shared taxonomy when new languages
reveal missing distinctions. Human calibration in the remaining evaluated
languages could also clarify how taxonomy resolution affects
reference-matching conclusions.

\section*{Limitations}
\label{sec:limitations}
The training corpus is proprietary and cannot be released, so exact reproduction of training remains unavailable. To support reproducibility, we release the METC annotation prompt, fine-tuned checkpoint, inference and evaluation code, training configuration, training-data statistics, taxonomy definitions, cross-taxonomy mappings, judge prompts, and frozen evaluation artifacts, including benchmark text and system, judge, and human annotations.

Reference-matching depends on hand-built mappings from each system's native
taxonomy into the corpus-native taxonomy. Adding a new system or corpus to the
comparison requires extending these mappings. The reported comparisons use one
fixed mapping for each system--corpus pair. These mappings are necessary for
classification, but they can collapse native distinctions, especially when METC emits a broader
operational label or a Compound label that the reference taxonomy does not
represent directly.
The main acceptability results are point estimates from one LLM judge. A full
repeat with a judge from another model family preserves the broad
observations but changes some pairwise orderings
(\S\ref{sec:results-judge-robustness}). Both judges are LLMs, the primary judge
shares a model family with METC's teacher, and verdicts remain sensitive to
the English descriptions used to render native labels. Hiding system names,
raw labels, and taxonomy prefixes reduces these effects but cannot eliminate
them. The cross-family judge evaluation and human calibration do not establish the reliability of teacher-generated training labels across languages.

The human evaluation covers English and German, with two independent verdicts
per item; the English study used a pool of twelve annotators. We do not have
human calibration for Czech, Korean, Ukrainian, or Chinese. Although METC is trained on data spanning over 200 languages, our benchmark evaluation covers six languages and does not establish comparable annotation quality beyond them. METC's 42-label taxonomy is designed as a shared
operational taxonomy rather than a comprehensive classification
of error types across all languages. Its design builds on ERRANT and therefore
retains English and Indo-European grammatical categories
and error conventions alongside cross-linguistic
adjustments. Consequently, language-specific grammatical
or morphological distinctions may be collapsed into
broader categories or require future taxonomy extensions.

\section*{Ethical Considerations}
METC is intended for error analysis and assistive annotation, not as an
independently validated pedagogical or high-stakes assessment system.
Incorrect labels could provide misleading feedback, particularly where the
shared taxonomy omits distinctions that matter for a language or domain.
Applications that expose labels to learners or use them in consequential
decisions should validate the relevant language and domain and retain human
review. The released evaluation artifacts include corpus text and annotations under the original dataset terms; human-evaluation records exclude annotator identities and free-form comments.

The human annotators were hourly contractors recruited through the public
job platforms Indeed and WayUp. They were paid at rates consistent with
professional annotation work, which are reviewed annually to account for
inflation and local cost of living. All annotators had at least two years of
editing experience and demonstrated strong written communication skills and
English proficiency through the Duolingo English Test or an equivalent
assessment. Professional linguistic training was not required. For
multilingual tasks, proficiency in the relevant non-English language was
verified through academic transcripts or writing samples.
The broader English annotation team included contractors residing in the
United States, South Africa, the United Kingdom, Canada, Belgium, and India.
The two German annotators resided in the United States.

\bibliography{custom}

\appendix

\section{Correspondence to ERRANT}
\label{app:errant-metc}

Table~\ref{tab:errant-metc} maps every ERRANT category to its METC
counterpart, grouped by the kind of relationship.

\begin{table*}[t]
\centering
\small
\resizebox{\textwidth}{!}{%
\begin{tabular}{@{}lll@{}}
\toprule
\textbf{ERRANT category} & \textbf{METC label $\rightarrow$ sublabel} & \textbf{Relationship} \\
\midrule
\multicolumn{3}{@{}l}{\textit{Preserved (category kept, sometimes generalized)}} \\
\addlinespace[2pt]
\texttt{ADJ} & Grammar $\rightarrow$ Adjective & preserved \\
\texttt{ADV} & Grammar $\rightarrow$ Adverb & preserved \\
\texttt{CONJ} & Grammar $\rightarrow$ Conjunction & preserved \\
\texttt{DET} & Grammar $\rightarrow$ Determiner & preserved \\
\texttt{NOUN} & Grammar $\rightarrow$ Noun & preserved \\
\texttt{PREP} & Grammar $\rightarrow$ Preposition & preserved \\
\texttt{PRON} & Grammar $\rightarrow$ Pronoun & preserved \\
\texttt{VERB} & Grammar $\rightarrow$ Verb & preserved \\
\texttt{MORPH} & Grammar $\rightarrow$ Morphology & preserved \\
\texttt{VERB:SVA} & Grammar $\rightarrow$ VerbSVA & preserved \\
\texttt{WO} & SentenceStructure $\rightarrow$ WordOrder & preserved \\
\texttt{PART} & Grammar $\rightarrow$ Particle & preserved, generalized \\
\midrule
\multicolumn{3}{@{}l}{\textit{Merged / redistributed (finer ERRANT subtypes combined)}} \\
\addlinespace[2pt]
\texttt{VERB:FORM}, \texttt{VERB:INFL}, \texttt{VERB:TENSE} & Grammar $\rightarrow$ VerbTenseForm & merged (3 $\rightarrow$ 1) \\
\texttt{ADJ:FORM} & Grammar $\rightarrow$ Adjective & folded in \\
\texttt{NOUN:NUM}, \texttt{NOUN:INFL}, \texttt{NOUN:POSS} & Grammar $\rightarrow$ Noun\,/\,Agreement & merged, redistributed \\
\midrule
\multicolumn{3}{@{}l}{\textit{Refined / split (coarse ERRANT type subdivided)}} \\
\addlinespace[2pt]
\texttt{PUNCT} & Punctuation $\rightarrow$ ClosingPunctuation\,/\,SentenceSplit\,/\,Other & refined (1 $\rightarrow$ 3) \\
\texttt{SPELL} & Spelling $\rightarrow$ Misspelling\,/\,SpellConfused\,/\,Diacritic & refined \\
\texttt{ORTH} (casing + whitespace) & Spelling (casing, Merge) \textit{and} Formatting (spacing) & split functionally \\
\texttt{CONTR} & Spelling $\rightarrow$ SpellConfused\_Confused & folded in \\
\midrule
\multicolumn{3}{@{}l}{\textit{Separated (ERRANT catch-all divided into structured branches)}} \\
\addlinespace[2pt]
\texttt{OTHER} & Rephrasing $\rightarrow$ Wordiness\,/\,Formality\,/\,Tone\,/\,Other & separated \\
\texttt{OTHER} (multi-type edits) & Other $\rightarrow$ Compound\_\{Trivial,Moderate,Complex\} & separated, graded \\
\midrule
\multicolumn{3}{@{}l}{\textit{Added (no ERRANT counterpart)}} \\
\addlinespace[2pt]
-- & Grammar $\rightarrow$ Agreement & added \\
-- & Grammar $\rightarrow$ Negation & added \\
-- & SentenceStructure $\rightarrow$ SentenceFragment & added \\
-- & Spelling $\rightarrow$ Diacritic & added (multilingual) \\
-- & Formatting $\rightarrow$ Date/Time/Currency/Measurement/SpellOutNumber & added \\
\midrule
\multicolumn{3}{@{}l}{\textit{Dropped}} \\
\addlinespace[2pt]
\texttt{UNK} & -- & dropped (corrections given) \\
\texttt{M:} / \texttt{R:} / \texttt{U:} operation prefixes & -- & dropped (recoverable from edit) \\
\bottomrule
\end{tabular}}
\caption{Correspondence between the ERRANT error-type framework
\citep{bryant-etal-2017-automatic} and the METC taxonomy. We preserve
ERRANT's part-of-speech-anchored categories, merge its fine morphological
subtypes, refine categories it bundles (notably \texttt{PUNCT},
\texttt{SPELL}, and \texttt{ORTH}), separate its catch-all \texttt{OTHER} into
structured Rephrasing and Compound branches, add categories with no ERRANT
counterpart, and drop the operation prefixes (recoverable from the edit) and
\texttt{UNK}.}
\label{tab:errant-metc}
\end{table*}

\section{Training hyperparameters}
\label{app:hyperparams}

Table~\ref{tab:hyperparams} lists the fine-tuning configuration of the student
model used in the experiments. The selected weights come from one fine-tuning
run with seed 42.

\begin{table}[h]
\centering
\footnotesize
\resizebox{\columnwidth}{!}{%
\begin{tabular}{ll}
\toprule
Parameter & Value \\
\midrule
Base encoder & \texttt{jhu-clsp/mmBERT-small} \\
Model class & \texttt{ModernBertForTokenClassification} \\
Classification outputs & 42 \\
Parameters & 140,658,474 \\
Maximum input length & 256 tokens \\
Marker tokens & \texttt{[EDIT]}, \texttt{[/EDIT]} \\
Optimizer & AdamW \\
Learning rate & 2e-5 \\
Batch size & 16 (train), 64 (eval) \\
Training stop & step 26,000 (epoch 2.176) \\
Warmup ratio & 0.1 \\
Weight decay & 0.01 \\
Precision & fp16 \\
Train / validation size & 1,529,258 / 169,918 \\
Validation data & 10\% internal held-out split \\
Selection metric & validation macro-F1 \\
Selected step & 23,000; macro-F1 0.7583 \\
Early stopping & patience of 3 evaluations \\
Seed & 42 \\
\bottomrule
\end{tabular}
}
\caption{Student fine-tuning hyperparameters.}
\label{tab:hyperparams}
\end{table}

\section{Language coverage}
\label{app:language_coverage}

The released METC checkpoint was trained on 1,529,258
examples spanning 208 distinct language codes, including
140 with at least 100 training examples and 71 with at
least 1,000 examples.

As of July 2026, versions of METC had been used as part
of Grammarly's grammatical error correction pipeline
across 24 languages: Czech, Danish, Dutch, English,
Finnish, French, German, Hindi, Hungarian, Indonesian,
Italian, Japanese, Korean, Norwegian, Polish, Portuguese,
Romanian, Slovak, Spanish, Swedish, Tagalog, Turkish,
Ukrainian, and Vietnamese.

Table~\ref{tab:language_coverage} lists all 208 language
codes grouped by the number of training examples in the
released checkpoint's training split.

\begin{table*}[t]
\centering
\small
\begin{tabular}{@{}p{0.79\textwidth}p{0.16\textwidth}@{}}
\toprule
\textbf{Languages (codes)} & \textbf{Training examples per code} \\
\midrule

English (en)
& 300k--500k \\

\addlinespace[2pt]

Portuguese (pt), French (fr), Italian (it), German (de),
Spanish (es)
& 50k--100k \\

\addlinespace[2pt]

Romanian (ro), Tagalog (tl), Hebrew (he), Swedish (sv),
Bosnian (bs), Korean (ko), Indonesian (id), Arabic (ar),
Persian (fa), Malay (ms), Ukrainian (uk), Polish (pl),
Chinese (zh), Turkish (tr), Afrikaans (af), Dutch (nl),
Russian (ru), Finnish (fi), Vietnamese (vi), Croatian (hr)
& 20k--50k \\

\addlinespace[2pt]

Georgian (ka), Czech (cs), Swahili (sw), Norwegian (no),
Danish (da), Mongolian (mn), Cebuano (ceb), Albanian (sq)
& 10k--20k \\

\addlinespace[2pt]

Sinhala (si), Thai (th), Japanese (ja), Azerbaijani (az),
Slovak (sk), Serbian (sr), Macedonian (mk), Uzbek (uz),
Bangla (bn), Hindi (hi), Kazakh (kk), Marathi (mr)
& 5k--10k \\

\addlinespace[2pt]

Hungarian (hu), Malayalam (ml), Gujarati (gu), Greek (el),
Tamil (ta), Estonian (et), Slovenian (sl), Waray (war),
Esperanto (eo), Somali (so), Catalan (ca), Tajik (tg),
Lao (lo), Lithuanian (lt), Haitian Creole (hat),
Kannada (kn), Bulgarian (bg), Central Kurdish (ckb),
Standard Latvian (lvs), Goan Konkani (gom),
M\={a}ori (mri), Javanese (jv), Standard Arabic (arb),
Iloko (ilo), Kinyarwanda (rw)
& 1k--5k \\

\addlinespace[2pt]

Persian (Afghanistan) (prs), Amharic (am), Yoruba (yo),
Hausa (ha), Lombard (lmo), Telugu (te), Icelandic (is),
Armenian (hy), Southern Pashto (pbt), Asturian (ast),
Urdu (ur), Kyrgyz (ky), Khmer (km), Galician (gl),
Eastern Yiddish (ydd), Nepali (individual language) (npi),
Samoan (smo)
& 500--1k \\

\addlinespace[2pt]

Sundanese (su), Northern Kurdish (kmr),
Plateau Malagasy (plt), Basque (eu), Tswana (tsn),
Zulu (zu), Southwestern Dinka (dik), Southern Sotho (sot),
Norwegian Nynorsk (nn), Belarusian (be), Xhosa (xho),
West Central Oromo (gaz), Occitan (oc), Papiamento (pap),
Mossi (mos), Shona (sn), Tsonga (tso),
Nigerian Fulfulde (fuv), Luxembourgish (ltz), Twi (twi),
Karachay-Balkar (krc), Northern Sotho (nso), Ganda (lug),
Fijian (fij), Faroese (fao), Wolof (wol),
Luo (Kenya and Tanzania) (luo), Irish (ga), Mizo (lus),
Pangasinan (pag), Nyanja (nya), Welsh (cy), Igbo (ibo),
Kimbundu (kmb), Central Kanuri (knc), Sindhi (sd),
South Azerbaijani (azb), Scottish Gaelic (gla), Fon (fon),
Burmese (my), Banjar (bjn), Assamese (as),
Norwegian Bokm\aa{}l (nb), Ligurian (lij),
Limburgish (lim), Turkmen (tuk), Punjabi (pa),
Dimli (individual language) (diq), Acehnese (ace),
Ewe (ewe), Ayacucho Quechua (quy), Rundi (run)
& 100--500 \\

\addlinespace[2pt]

Bambara (bam), Maltese (mt), Cameroon Pidgin (wes),
Bemba (bem), Sango (sag), Odia (individual language) (ory),
Sanskrit (sa), Tahitian (tah), Minangkabau (min),
Sicilian (scn), Umbundu (umb), Tigrinya (ti), Lingala (lin),
Kachin (kac), Nigerian Pidgin (pcm), Bislama (bis),
Romansh (roh), Kabyle (kab), Tok Pisin (tpi),
Buginese (bug), Latgalian (ltg), Tatar (tt), Guarani (grn),
Akan (aka), Tibetan (bo), Dyula (dyu), Venetian (vec),
Kamba (kam), Luba-Lulua (lua), Swati (ssw),
Central Aymara (ayr), Kabuverdianu (kea), Nias (nia),
Tumbuka (tum), Balinese (ban), Maithili (mai),
Kikuyu (kik), Tamasheq (taq), Nuer (nus), Uyghur (ug),
Adyghe (ady), Crimean Tatar (crh), Tongan (ton),
Chuvash (chv), Chokwe (cjk), Kongo (kon), Chechen (che),
Kashmiri (kas), Kalaallisut (kal), Navajo (nav),
Mingrelian (xmf), Friulian (fur), Bashkir (bak),
Russia Buriat (bxr), Silesian (szl), Mapuche (arn)
& 10--100 \\

\addlinespace[2pt]

Ossetic (oss), Ewondo (ewo), Kabiy\`{e} (kbp),
Manipuri (mni), Sardinian (srd), Abkhazian (abk),
Bhojpuri (bho), Awadhi (awa), Chhattisgarhi (hne),
Dzongkha (dzo), Magahi (mag), Shan (shn)
& 1--10 \\

\bottomrule
\end{tabular}
\caption{Training-data coverage by language code. Each range
includes its lower bound and excludes its upper bound;
k denotes one thousand examples. Language identifiers are
reported as recorded in the training data.}
\label{tab:language_coverage}
\end{table*}

\section{LLM-as-judge prompts}
\label{app:prompts}

The acceptability judge described in \S\ref{sec:eval-judge} sees the source
sentence, the corrected sentence, the edit rendered as a \texttt{'src' => 'tgt'}
diff, and a single-sentence English description of the system's label. The judge never
receives the system identifier, the native label code, the taxonomy prefix, or
the M/U/R operation tag. Each judge call returns strict JSON; per-axis
rationales are required to discourage axis-averaging.

\begin{small}
\begin{verbatim}
You are a senior linguist evaluating one
error-correction edit.

Language: {language}
Source sentence: "{source}"
Corrected sentence: "{target}"
The edit being judged: {edit_diff}
The classifier described this edit as:
{label_description}

Rate the classifier on two independent axes.
For each, choose ONE of:
  Good        - fully appropriate
  Acceptable  - defensible but not the best
                choice (span is not semantically
                clean, or a coarser super-type
                description)
  Bad         - wrong or misleading

span_quality:
  Good       = the span exactly covers the
               error, no more no less
  Acceptable = the span covers the error but is
               semantically confusing or not
               precise
  Bad        = the span includes unrelated text
               that makes it unclear what error
               the edit is fixing

label_quality:
  Good       = the description is the most
               precise correct interpretation
               of this edit
  Acceptable = the description is a coarser
               but still valid interpretation
  Bad        = the description is wrong,
               irrelevant, or contradicts what
               the edit does

Return only a JSON object, no prose, no
markdown:
{
  "span_quality":  "Good" | "Acceptable" | "Bad",
  "label_quality": "Good" | "Acceptable" | "Bad",
  "span_reason":   "<one short sentence
                    citing the source span>",
  "label_reason":  "<one short sentence citing
                    why the description fits
                    or does not>"
}
\end{verbatim}
\end{small}

\paragraph{Label description renderer.}
Each system's native label is converted to a single declarative English
sentence before it reaches the judge. The renderer routes by label shape:
METC labels are keyed on \texttt{group/sublabel}; CoNLL, MERLIN, and UA-GEC
reference labels are keyed through taxonomy-specific dictionaries; ERRANT-family
labels are parsed as \texttt{M/U/R:CAT[:SUB]}; and KAGAS labels are rendered
from their POS-pattern structure: single categories use the per-category
renderings of Table~\ref{tab:errant-desc}, and transition labels
(\texttt{R:P1 -> P2}, the scheme's replacement pattern) render both sides
faithfully, so \texttt{R:AUX VERB -> VERB AUX} reads ``Problem with an
auxiliary verb and a verb corrected to a verb and an auxiliary verb.''
The operation tag is intentionally suppressed for ERRANT-family labels when it
is already visible in the edit diff, because pilot runs showed that it could
draw attention away from the label category. Across systems, each label is
rendered as the same kind of neutral English statement so the judge sees
comparable phrasing rather than raw taxonomy strings.
For example, METC \texttt{Grammar/VerbTenseForm} renders as ``Problem with
verb tense or form.'', ERRANT \texttt{R:VERB:TENSE} renders as ``Problem with
a verb tense.'', and METC \texttt{Spelling/Misspelling} renders as ``Problem
with the spelling.'' KAGAS labels for Korean are handled with a language-aware POS
override, so UD \texttt{ADP} is rendered as a postposition rather than a
preposition.

\section{Exact-span sensitivity}
\label{app:exact-span}

The reference-matching results in Table~\ref{tab:exp-h} use one-to-one overlap
matching because annotation
systems can segment the same correction differently. Table~\ref{tab:exact-span}
repeats the analysis while requiring exact source-span equality. Absolute
scores fall, particularly for METC on English and Chinese, but every
classification ordering is preserved. Detection orderings are also preserved
except that the two German baselines exchange positions and jp\_errant moves
from statistically indistinguishable from METC to higher on Chinese. METC
remains closer than the German ERRANT modification on German and jp\_errant on
Czech. The main reference-matching observations therefore do not depend on
relaxed span matching.

\begin{table*}[t]
\centering
\small
\setlength{\tabcolsep}{2.5pt}
\begin{tabular}{lrrrrrrr}
\toprule
System & EN0 & EN1 & DE & CS & KO & UK & ZH \\
\midrule
\multicolumn{8}{l}{\textit{Detection $F_{0.5}$}} \\
METC       & 0.621\,$\pm$\,0.025 & 0.645\,$\pm$\,0.020 & 0.486\,$\pm$\,0.032 & 0.600\,$\pm$\,0.012 & 0.923\,$\pm$\,0.019 & 0.661\,$\pm$\,0.017 & 0.636\,$\pm$\,0.017 \\
ERRANT     & 0.825\,$\pm$\,0.020 & 0.860\,$\pm$\,0.018 & -- & -- & -- & -- & -- \\
gERRANT    & -- & -- & 0.476\,$\pm$\,0.029 & -- & -- & -- & -- \\
jp\_errant & 0.816\,$\pm$\,0.019 & 0.844\,$\pm$\,0.019 & 0.470\,$\pm$\,0.029 & 0.569\,$\pm$\,0.012 & 0.999\,$\pm$\,0.001 & 0.865\,$\pm$\,0.013 & 0.727\,$\pm$\,0.015 \\
\midrule
\multicolumn{8}{l}{\textit{Classification $F_{0.5}$}} \\
METC         & 0.406\,$\pm$\,0.022 & 0.454\,$\pm$\,0.019 & 0.297\,$\pm$\,0.024 & 0.418\,$\pm$\,0.011 & 0.332\,$\pm$\,0.022 & 0.409\,$\pm$\,0.018 & 0.469\,$\pm$\,0.017 \\
METC-teacher & 0.389\,$\pm$\,0.021 & 0.436\,$\pm$\,0.018 & 0.306\,$\pm$\,0.023 & 0.422\,$\pm$\,0.011 & 0.330\,$\pm$\,0.022 & 0.447\,$\pm$\,0.019 & 0.477\,$\pm$\,0.018 \\
ERRANT       & 0.564\,$\pm$\,0.022 & 0.634\,$\pm$\,0.022 & -- & -- & -- & -- & -- \\
gERRANT      & -- & -- & 0.261\,$\pm$\,0.021 & -- & -- & -- & -- \\
jp\_errant   & 0.521\,$\pm$\,0.022 & 0.572\,$\pm$\,0.020 & 0.130\,$\pm$\,0.014 & 0.136\,$\pm$\,0.006 & 0.999\,$\pm$\,0.001 & 0.520\,$\pm$\,0.015 & 0.425\,$\pm$\,0.017 \\
\bottomrule
\end{tabular}
\caption{Exact-span sensitivity analysis. Values are $F_{0.5}$ with 95\%
paired sentence-bootstrap intervals shown as $\pm$ half-width. METC-teacher
shares METC's extractor, so its detection row is omitted. EN0 and EN1 are the
two CoNLL-2014 annotators.}
\label{tab:exact-span}
\end{table*}

\section{Label descriptions for the acceptability judge}
\label{app:descriptions}

The acceptability judge (\S\ref{sec:eval-judge}) sees each system's native label as a single English description, with no system name or raw label code. This appendix lists those descriptions in full: Table~\ref{tab:metc-desc} for the 42 METC labels, Table~\ref{tab:errant-desc} for the per-category renderings shared by the ERRANT-family and KAGAS baselines, and Tables~\ref{tab:ref-desc}--\ref{tab:ref-desc-uagec} for the reference-corpus labels.

\begin{table*}[p]
\centering
\scriptsize
\renewcommand{\arraystretch}{0.92}
\setlength{\tabcolsep}{3pt}
\begin{tabular}{@{}p{0.31\textwidth} p{0.64\textwidth}@{}}
\toprule \textbf{METC label} & \textbf{Description} \\ \midrule
Formatting/AbbreviationFormat & Problem with how an abbreviation is formatted. \\
Formatting/CurrencyFormat & Problem with how a currency amount is formatted. \\
Formatting/DateFormat & Problem with how a date is formatted. \\
Formatting/MeasurementFormat & Problem with how a measurement is formatted. \\
Formatting/MissingSpace & Problem with a missing space. \\
Formatting/Other & Problem with formatting. \\
Formatting/RedundantSpace & Problem with an extra space. \\
Formatting/SpellOutNumber & Problem with whether a number is written as a numeral or spelled out. \\
Formatting/TimeFormat & Problem with how a time expression is formatted. \\
Grammar/Adjective & Problem with an adjective. \\
Grammar/Adverb & Problem with an adverb. \\
Grammar/Agreement & Problem with grammatical agreement (number, person, or gender between a noun and its determiner, adjective, pronoun, or quantifier; not subject-verb agreement). \\
Grammar/Conjunction & Problem with a conjunction. \\
Grammar/Determiner & Problem with an article or determiner. \\
Grammar/Morphology & Problem with a word form. \\
Grammar/Negation & Problem with negation. \\
Grammar/Noun & Problem with a noun. \\
Grammar/Particle & Problem with a particle. \\
Grammar/Preposition & Problem with a preposition. \\
Grammar/Pronoun & Problem with a pronoun. \\
Grammar/Verb & Problem with a verb (lexical choice, missing or extra auxiliary, or the to-infinitive marker; not tense or subject-verb agreement). \\
Grammar/VerbSVA & Problem with subject-verb agreement. \\
Grammar/VerbTenseForm & Problem with verb tense or form. \\
Other/Compound\_Complex & Problem with an extensive, multi-part correction. \\
Other/Compound\_Moderate & Problem with a moderately complex, multi-part correction. \\
Other/Compound\_Trivial & Problem with a small, multi-part correction. \\
Punctuation/ClosingPunctuation & Problem with sentence-ending punctuation. \\
Punctuation/Other & Problem with internal punctuation. \\
Punctuation/SentenceSplit & Problem with how sentences are split or joined. \\
Rephrasing/Formality & Problem with the formality of the wording. \\
Rephrasing/Other & Problem with the clarity of the wording. \\
Rephrasing/Tone & Problem with the tone of the wording. \\
Rephrasing/Wordiness & Problem with wordiness. \\
SentenceStructure/SentenceFragment & Problem with a sentence missing a required subject or verb. \\
SentenceStructure/WordOrder & Problem with word order. \\
Spelling/Diacritic & Problem with a diacritical mark (accent, umlaut, cedilla, etc.). \\
Spelling/Merge & Problem with two adjacent words that should be joined into one. \\
Spelling/Misspelling & Problem with the spelling. \\
Spelling/Other & Problem with a spelling issue. \\
Spelling/SpellConfused\_Capitalize & Problem with a word that should be capitalized. \\
Spelling/SpellConfused\_Confused & Problem with a correctly-spelled word used in place of another (e.g., were vs where). \\
Spelling/SpellConfused\_Lowercase & Problem with a word that should not be capitalized. \\
\bottomrule
\end{tabular}
\caption{One-sentence descriptions for the 42 METC labels.}
\label{tab:metc-desc}
\end{table*}

\begin{table*}[p]
\centering
\scriptsize
\renewcommand{\arraystretch}{0.92}
\setlength{\tabcolsep}{3pt}
\begin{tabular}{@{}p{0.25\textwidth} p{0.70\textwidth}@{}}
\toprule \textbf{Category} & \textbf{Rendered description} \\ \midrule
ADJ & Problem with an adjective. \\
ADJ:FORM & Problem with an adjective form. \\
ADP & Problem with a preposition. \\
ADV & Problem with an adverb. \\
AUX & Problem with an auxiliary verb. \\
AUX:FORM & Problem with an auxiliary verb form. \\
CCONJ & Problem with a coordinating conjunction. \\
CO & Problem with the order of characters within a word. \\
CONJ & Problem with a conjunction. \\
CONTR & Problem with a contraction. \\
DE & Problem with the structural particle \textit{de} (three homophonous particles). \\
DET & Problem with an article or determiner. \\
DET:FORM & Problem with an article or determiner form. \\
INTJ & Problem with an interjection. \\
MC & Problem with a missing sentence component. \\
MORPH & Problem with a word form. \\
MULTI & Problem with a misspelled character similar in both pronunciation and shape. \\
NOUN & Problem with a noun. \\
NOUN:FORM & Problem with a noun form. \\
NOUN:INFL & Problem with a noun inflection. \\
NOUN:NUM & Problem with a noun's number. \\
NOUN:POSS & Problem with a noun's possessive form. \\
NUM & Problem with a number. \\
ORTH & Problem with the capitalization or spacing. \\
PART & Problem with a particle. \\
PINYIN & Problem with a misspelled character with a similar pronunciation. \\
PNOUN & Problem with a proper noun. \\
PREP & Problem with a preposition. \\
PRON & Problem with a pronoun. \\
PRON:FORM & Problem with a pronoun form. \\
PROPN & Problem with a proper noun. \\
PUNCT & Problem with a punctuation mark. \\
QUAN & Problem with a measure word or quantifier. \\
SCONJ & Problem with a subordinating conjunction. \\
SHAPE & Problem with a misspelled character with a visually similar shape. \\
SPELL & Problem with the spelling. \\
SPELL:PUNCT & Problem with a punctuation mark. \\
SYM & Problem with a symbol. \\
VERB (English) & Problem with a verb (its lexical choice; not tense, form, subject-verb agreement, or inflection). \\
VERB (other languages) & Problem with a verb. \\
VERB:FORM & Problem with a verb form. \\
VERB:INFL & Problem with a verb inflection. \\
VERB:SVA & Problem with subject-verb agreement. \\
VERB:TENSE & Problem with a verb tense. \\
X & Problem with a token. \\
WO & Problem with word order. \\
OTHER & Problem with a word or phrase. \\
UNK & Problem with a word or phrase. \\
\bottomrule
\end{tabular}
\caption{Per-category renderings for ERRANT-family and KAGAS labels.}
\label{tab:errant-desc}
\end{table*}

\begin{table*}[p]
\centering
\scriptsize
\renewcommand{\arraystretch}{0.92}
\setlength{\tabcolsep}{3pt}
\begin{tabular}{@{}p{0.13\textwidth} p{0.26\textwidth} p{0.56\textwidth}@{}}
\toprule \textbf{Taxonomy} & \textbf{Label} & \textbf{Description} \\ \midrule
CoNLL & Vt & Problem with verb tense. \\
 & Vm & Problem with a modal verb. \\
 & V0 & Problem with a missing verb. \\
 & Vform & Problem with a verb form. \\
 & SVA & Problem with subject-verb agreement. \\
 & ArtOrDet & Problem with an article or determiner. \\
 & Nn & Problem with a noun's number (singular/plural). \\
 & Npos & Problem with a noun's possessive form. \\
 & Pform & Problem with a pronoun's form. \\
 & Pref & Problem with a pronoun reference. \\
 & Prep & Problem with a preposition. \\
 & Wci & Problem with word choice. \\
 & Wa & Problem with how an acronym is used. \\
 & Wform & Problem with a word's form within the same word family. \\
 & Wtone & Problem with the tone of the wording. \\
 & Srun & Problem with sentences run together. \\
 & Smod & Problem with a modifier (refers to the wrong word). \\
 & Spar & Problem with parallelism in a list. \\
 & Sfrag & Problem with an incomplete sentence. \\
 & Ssub & Problem with a subordinate clause. \\
 & WOinc & Problem with word order. \\
 & WOadv & Problem with the position of an adverb or adjective. \\
 & Trans & Problem with a linking word. \\
 & Mec & Problem with punctuation, capitalization, or spelling. \\
 & Rloc- & Problem with a redundant word or phrase. \\
 & Cit & Problem with a citation. \\
 & Others & Problem with a word or phrase. \\
 & Um & Problem with unclear meaning. \\
\bottomrule
\end{tabular}
\caption{Descriptions for the CoNLL-2014 reference labels.}
\label{tab:ref-desc}
\end{table*}

\begin{table*}[p]
\centering
\scriptsize
\renewcommand{\arraystretch}{0.92}
\setlength{\tabcolsep}{3pt}
\begin{tabular}{@{}p{0.13\textwidth} p{0.26\textwidth} p{0.56\textwidth}@{}}
\toprule \textbf{Taxonomy} & \textbf{Label} & \textbf{Description} \\ \midrule
MERLIN & C\_Coh\_jump & Problem with a gap in the flow of information. \\
 & C\_Coh\_ref & Problem with what a word refers back to. \\
 & C\_Coh\_txtstruct & Problem with how the parts of the text connect. \\
 & C\_Con\_accur & Problem with content accuracy. \\
 & G\_Agr & Problem with grammatical agreement. \\
 & G\_Art & Problem with an article. \\
 & G\_Clit & Problem with a short unstressed pronoun particle. \\
 & G\_Conj & Problem with a conjunction. \\
 & G\_Inflect\_Inexist & Problem with a non-existent word form. \\
 & G\_Inflect\_Inexist/adj & Problem with a non-existent adjective form. \\
 & G\_Inflect\_Inexist/noun & Problem with a non-existent noun form. \\
 & G\_Inflect\_Inexist/verb & Problem with a non-existent verb form. \\
 & G\_Morphol\_Wrong & Problem with word inflection. \\
 & G\_Morphol\_Wrong/case & Problem with grammatical case. \\
 & G\_Morphol\_Wrong/gend & Problem with grammatical gender. \\
 & G\_Morphol\_Wrong/numb & Problem with grammatical number. \\
 & G\_Neg & Problem with negation. \\
 & G\_Pos & Problem with the part of speech. \\
 & G\_Prep & Problem with a preposition. \\
 & G\_Refl & Problem with a reflexive form. \\
 & G\_Refl/pronrefl & Problem with a reflexive pronoun. \\
 & G\_Refl/pronreflposs & Problem with a reflexive possessive pronoun. \\
 & G\_Valency & Problem with a verb's object or complement. \\
 & G\_Verb & Problem with verb tense, voice, mood, or aspect. \\
 & G\_Verb/asp & Problem with verb aspect. \\
 & G\_Verb/md & Problem with verb mood. \\
 & G\_Verb/tns & Problem with verb tense. \\
 & G\_Verb/vc & Problem with verb voice (active vs passive). \\
 & G\_Verb\_compl & Problem with a complex verb construction (auxiliary, participle, or infinitive). \\
 & G\_Verb\_main & Problem with the main verb. \\
 & G\_Wo & Problem with word order. \\
 & H\_Intellts & Problem with sentence-level intelligibility. \\
 & H\_Intelltxt & Problem with text intelligibility. \\
 & O\_Abbrev & Problem with an abbreviation. \\
 & O\_Apostr & Problem with an apostrophe. \\
 & O\_Capit & Problem with capitalization. \\
 & O\_Graph & Problem with the spelling. \\
 & O\_Graph/act & Problem with a diacritical mark (accent, umlaut, cedilla, etc.). \\
 & O\_Graph/trans & Problem with the order of letters within a word. \\
 & O\_Punct & Problem with punctuation. \\
 & O\_Wordbd & Problem with word boundaries (joining or splitting). \\
 & P\_Pol\_dir & Problem with politeness or directness. \\
 & P\_Request & Problem with a request. \\
 & S\_Form & Problem with politeness form. \\
 & S\_Txt & Problem with text style. \\
 & S\_Txt/grfw & Problem with a greeting or sign-off phrase. \\
 & S\_Txt/opcl & Problem with how the text opens or closes. \\
 & S\_Var & Problem with register. \\
 & V\_FS & Problem with a set phrase. \\
 & V\_FS\_form & Problem with the form of a set phrase. \\
 & V\_Sequence\_lexgrammer\_inc & Problem with a multi-part correction spanning more than one kind of change. \\
 & V\_Wordform & Problem with word form. \\
 & V\_form\_FS\_incompr & Problem with an unclear set phrase. \\
 & V\_form\_word\_fs\_nonexist & Problem with a non-existent word or set phrase. \\
 & V\_semconn\_at\_word\_fs & Problem with the meaning connection of a word or set phrase. \\
 & V\_semdenot\_word\_fs & Problem with the meaning of a word or set phrase. \\
 & V\_semimprec & Problem with imprecise word choice. \\
\bottomrule
\end{tabular}
\caption{Descriptions for the MERLIN reference labels.}
\label{tab:ref-desc-merlin}
\end{table*}

\begin{table*}[p]
\centering
\scriptsize
\renewcommand{\arraystretch}{0.92}
\setlength{\tabcolsep}{3pt}
\begin{tabular}{@{}p{0.13\textwidth} p{0.26\textwidth} p{0.56\textwidth}@{}}
\toprule \textbf{Taxonomy} & \textbf{Label} & \textbf{Description} \\ \midrule
UA-GEC & Spelling & Problem with the spelling. \\
 & Punctuation & Problem with punctuation. \\
 & G/Case & Problem with grammatical case. \\
 & G/Number & Problem with grammatical number (singular/plural). \\
 & G/Gender & Problem with grammatical gender. \\
 & G/Tense & Problem with verb tense. \\
 & G/VerbVoice & Problem with verb voice (active vs passive). \\
 & G/PartVoice & Problem with participle voice. \\
 & G/VerbAForm & Problem with a verb's adjectival form. \\
 & G/Aspect & Problem with verb aspect (perfective vs imperfective). \\
 & G/Prep & Problem with a preposition. \\
 & G/Participle & Problem with a participle. \\
 & G/Conjunction & Problem with a conjunction. \\
 & G/Comparison & Problem with the form of comparison (comparative or superlative). \\
 & G/Particle & Problem with a particle. \\
 & G/UngrammaticalStructure & Problem with sentence structure (ungrammatical construction). \\
 & G/Other & Problem with a grammatical construction. \\
\bottomrule
\end{tabular}
\caption{Descriptions for the UA-GEC reference labels.}
\label{tab:ref-desc-uagec}
\end{table*}

\section{Mapping tables}
\label{app:mappings}

Each system's native labels are mapped into the corpus-native reference taxonomy
before classification is computed. Table~\ref{tab:mappings} records the fixed
mappings used for all reported classification results. These mappings are
evaluation artifacts only; they are not part of METC.
Table~\ref{tab:mappings} lists the METC source label for each of the five
reference target taxonomies.

\begin{table*}[t]
\centering
\footnotesize
\setlength{\tabcolsep}{4pt}
\resizebox{\textwidth}{!}{%
\begin{tabular}{l l l l l l}
\toprule
\textbf{METC label} & \textbf{CoNLL} & \textbf{MERLIN} & \textbf{KoLLA} & \textbf{UA-GEC} & \textbf{ChERRANT} \\
\midrule
\multicolumn{6}{l}{\textit{Formatting}} \\
Formatting/AbbreviationFormat   & Mec      & O\_Abbrev                 & R:SPELL & Punctuation & PUNCT \\
Formatting/CurrencyFormat       & Mec      & O\_Abbrev                 & R:SPELL & Punctuation & PUNCT \\
Formatting/DateFormat           & Mec      & O\_Abbrev                 & R:SPELL & Punctuation & PUNCT \\
Formatting/MeasurementFormat    & Mec      & O\_Abbrev                 & R:SPELL & Punctuation & PUNCT \\
Formatting/MissingSpace         & Mec      & O\_Wordbd                 & R:SPELL & Spelling & SPELL \\
Formatting/Other                & Mec      & O\_Abbrev                 & R:SPELL & Punctuation & PUNCT \\
Formatting/RedundantSpace       & Mec      & O\_Wordbd                 & R:SPELL & Spelling & SPELL \\
Formatting/SpellOutNumber       & Mec      & O\_Abbrev                 & R:SPELL & Spelling & NUM \\
Formatting/TimeFormat           & Mec      & O\_Abbrev                 & R:SPELL & Punctuation & PUNCT \\
\midrule
\multicolumn{6}{l}{\textit{Grammar}} \\
Grammar/Adjective               & Wci      & G\_Morphol\_Wrong         & R:ADJ   & G/Case & ADJ \\
Grammar/Adverb                  & Wform    & V\_Wordform               & R:ADV   & G/Other & ADV \\
Grammar/Agreement               & Nn       & G\_Agr                    & R:VERB  & G/Number & OTHER \\
Grammar/Conjunction             & Trans    & G\_Conj                   & R:CCONJ & G/Conjunction & CONJ \\
Grammar/Determiner              & ArtOrDet & G\_Art                    & R:DET   & G/Other & AUX \\
Grammar/Morphology              & Wform    & G\_Morphol\_Wrong         & R:NOUN  & G/Case & OTHER \\
Grammar/Negation                & Others   & G\_Neg                    & R:VERB  & G/Particle & ADV \\
Grammar/Noun                    & Nn       & G\_Morphol\_Wrong         & R:NOUN  & G/Case & NOUN \\
Grammar/Particle                & Prep     & G\_Verb\_compl            & R:PART  & G/Particle & AUX \\
Grammar/Preposition             & Prep     & G\_Prep                   & R:ADP   & G/Prep & PREP \\
Grammar/Pronoun                 & Pref     & G\_Morphol\_Wrong         & R:PRON  & G/Case & PRON \\
Grammar/Verb                    & Wci      & G\_Verb                   & R:VERB  & G/Tense & VERB \\
Grammar/VerbSVA                 & SVA      & G\_Agr                    & R:VERB  & G/Number & VERB \\
Grammar/VerbTenseForm           & Vt       & G\_Verb                   & R:VERB  & G/Tense & VERB \\
\midrule
\multicolumn{6}{l}{\textit{Other (Compound)}} \\
Other/Compound\_Trivial         & Others   & V\_Sequence\_lexgrammer\_inc & R:SPELL & G/UngrammaticalStructure & OTHER \\
Other/Compound\_Moderate        & Others   & V\_Sequence\_lexgrammer\_inc & R:SPELL & G/UngrammaticalStructure & OTHER \\
Other/Compound\_Complex         & Others   & V\_Sequence\_lexgrammer\_inc & R:SPELL & G/UngrammaticalStructure & OTHER \\
\midrule
\multicolumn{6}{l}{\textit{Punctuation}} \\
Punctuation/ClosingPunctuation  & Mec      & O\_Punct                  & R:PUNCT & Punctuation & PUNCT \\
Punctuation/Other               & Mec      & O\_Punct                  & R:PUNCT & Punctuation & PUNCT \\
Punctuation/SentenceSplit       & Mec      & O\_Punct                  & R:PUNCT & Punctuation & PUNCT \\
\midrule
\multicolumn{6}{l}{\textit{Rephrasing}} \\
Rephrasing/Formality            & Wtone    & S\_Var                    & R:VERB  & G/Other & OTHER \\
Rephrasing/Other                & Others   & V\_semdenot\_word\_fs     & R:NOUN  & G/Other & OTHER \\
Rephrasing/Tone                 & Wtone    & S\_Var                    & R:VERB  & G/Other & OTHER \\
Rephrasing/Wordiness            & Rloc-    & S\_Txt                    & R:NOUN  & G/UngrammaticalStructure & OTHER \\
\midrule
\multicolumn{6}{l}{\textit{SentenceStructure}} \\
SentenceStructure/SentenceFragment & Sfrag & H\_Intellts               & R:VERB  & G/UngrammaticalStructure & OTHER \\
SentenceStructure/WordOrder        & WOinc & G\_Wo                     & R:WO    & G/UngrammaticalStructure & W \\
\midrule
\multicolumn{6}{l}{\textit{Spelling}} \\
Spelling/Diacritic              & Mec      & O\_Graph                  & R:SPELL & Spelling & SPELL \\
Spelling/Merge                  & Mec      & O\_Wordbd                 & R:SPELL & Spelling & SPELL \\
Spelling/Misspelling            & Mec      & O\_Graph                  & R:SPELL & Spelling & SPELL \\
Spelling/Other                  & Mec      & O\_Graph                  & R:SPELL & Spelling & SPELL \\
Spelling/SpellConfused\_Capitalize & Mec   & O\_Capit                  & R:SPELL & Spelling & SPELL \\
Spelling/SpellConfused\_Confused   & Mec   & O\_Graph                  & R:SPELL & Spelling & SPELL \\
Spelling/SpellConfused\_Lowercase  & Mec   & O\_Capit                  & R:SPELL & Spelling & SPELL \\
\bottomrule
\end{tabular}
}
\caption{METC label routed into each reference taxonomy used in the
reference-matching protocol. KAGAS labels for KoLLA are shown without the token-pattern
body; the renderer collapses pattern bodies to the head category. The
three Compound levels share a single target per taxonomy because all reference
taxonomies are flat (no compound bucket), so granularity is preserved only in
METC's own predictions.}
\label{tab:mappings}
\end{table*}

The remaining reference-matching baselines include ERRANT and jp\_errant for English,
gERRANT and jp\_errant for German, and jp\_errant for Czech and
Ukrainian. Each is mapped into the corpus-native target by the
version-controlled reference-matching code. The lookup tables below document the
main mappings expanded in this appendix; the mapping code defines rows whose
native emission inventory is not expanded here. KoLLA has
no separate mapping for jp\_errant\_bea-ko because the BEA-2025 KAGAS variant
emits labels that are already in the KoLLA reference taxonomy; the corpus-native
reference label is consumed identity-mapped.

\paragraph{ERRANT $\to$ CoNLL.}
Twenty-five ERRANT base categories are routed to CoNLL via the flat lookup in
Table~\ref{tab:errant-conll}. CoNLL has dedicated classes for several specific
operation/category combinations (a missing verb is \texttt{V0}, a redundant
word is \texttt{Rloc-}, a wrong contraction reads as a register choice
\texttt{Wtone}); for those combinations the operation prefix changes the
semantic content of the edit, so the flat category lookup is supplemented by
the operation-prefix overrides in Table~\ref{tab:errant-prefix-overrides}.
Bare \texttt{M:/U:/R:} prefixes are stripped before category lookup when no
override applies.

\begin{table}[h]
\centering
\small
\setlength{\tabcolsep}{4pt}
\begin{tabular}{l l}
\toprule
\textbf{ERRANT category} & \textbf{CoNLL} \\
\midrule
ADJ          & Wform \\
ADJ:FORM     & Wform \\
ADV          & Wform \\
CONJ         & Trans \\
CONTR        & Vform \\
DET          & ArtOrDet \\
MORPH        & Wform \\
NOUN         & Wci \\
NOUN:INFL    & Nn \\
NOUN:NUM     & Nn \\
NOUN:POSS    & Npos \\
ORTH         & Mec \\
OTHER        & Others \\
PART         & Prep \\
PREP         & Prep \\
PRON         & Pref \\
PUNCT        & Mec \\
SPELL        & Mec \\
UNK          & Others \\
VERB         & Vform \\
VERB:FORM    & Vform \\
VERB:INFL    & Vform \\
VERB:SVA     & SVA \\
VERB:TENSE   & Vt \\
WO           & WOinc \\
\bottomrule
\end{tabular}
\caption{ERRANT $\to$ CoNLL category lookup. The M/U/R operation prefix is
stripped before lookup unless an entry in Table~\ref{tab:errant-prefix-overrides}
takes precedence.}
\label{tab:errant-conll}
\end{table}

\begin{table}[h]
\centering
\small
\setlength{\tabcolsep}{4pt}
\begin{tabular}{l l l}
\toprule
\textbf{(prefix, category)} & \textbf{CoNLL} & \textbf{Reading} \\
\midrule
M:VERB   & V0    & missing verb \\
R:VERB   & Wci   & verb word-choice \\
R:CONTR  & Wtone & formal/informal register \\
R:ADJ    & Wci   & adjective word-choice \\
R:ADV    & Wci   & adverb word-choice \\
U:ADJ    & Rloc- & redundant adjective \\
U:ADV    & Rloc- & redundant adverb \\
U:VERB   & Rloc- & redundant verb \\
U:OTHER  & Rloc- & redundant phrase \\
\bottomrule
\end{tabular}
\caption{ERRANT prefix-sensitive overrides applied before the category lookup
in Table~\ref{tab:errant-conll}. The third column gives the CoNLL semantic
reading that justifies the override.}
\label{tab:errant-prefix-overrides}
\end{table}

\paragraph{gERRANT $\to$ MERLIN.}
The German baseline taxonomy (Appendix~\ref{app:boyd-port}) is routed into
MERLIN through Table~\ref{tab:gerrant-merlin}. Boyd-style category prefixes
\texttt{M:/U:/R:} are stripped before lookup. The two MERLIN classes
\texttt{G\_Inflect\_Inexist} and \texttt{O\_Apostr} are reachable only through
gERRANT; METC's flat taxonomy collapses them into \texttt{G\_Morphol\_Wrong}
and \texttt{O\_Graph} respectively.

\begin{table}[h]
\centering
\small
\setlength{\tabcolsep}{4pt}
\begin{tabular}{l l}
\toprule
\textbf{gERRANT category} & \textbf{MERLIN} \\
\midrule
ADJ          & G\_Morphol\_Wrong \\
ADJ:FORM     & G\_Morphol\_Wrong \\
ADV          & V\_Wordform \\
CONJ / SCONJ / CCONJ & G\_Conj \\
CONTR        & O\_Apostr \\
DET          & G\_Art \\
MORPH        & G\_Pos \\
NOUN         & G\_Morphol\_Wrong \\
NOUN:FORM    & G\_Morphol\_Wrong \\
NOUN:NUM     & G\_Morphol\_Wrong \\
NOUN:INFL    & G\_Inflect\_Inexist \\
NOUN:POSS    & G\_Refl \\
ORTH         & O\_Capit \\
OTHER        & V\_semdenot\_word\_fs \\
PART         & G\_Verb\_compl \\
PREP / ADP   & G\_Prep \\
PRON         & G\_Refl \\
PROPN / PNOUN & G\_Morphol\_Wrong \\
PUNCT        & O\_Punct \\
SPELL        & O\_Graph \\
VERB         & G\_Verb \\
VERB:FORM    & G\_Verb\_compl \\
VERB:TENSE   & G\_Verb \\
VERB:SVA     & G\_Agr \\
VERB:INFL    & G\_Inflect\_Inexist \\
WO           & G\_Wo \\
NUM          & O\_Abbrev \\
INTJ / X / UNK & V\_semdenot\_word\_fs \\
SYM          & O\_Abbrev \\
\bottomrule
\end{tabular}
\caption{gERRANT $\to$ MERLIN category lookup.}
\label{tab:gerrant-merlin}
\end{table}

\paragraph{jp\_errant\_bea $\to$ MERLIN and UA-GEC.}
The multilingual ERRANT modification \citep{qiu-etal-2025-multilingual} emits patterns
of the form \texttt{M/U/R:UPOS} (optionally with a pattern body
\texttt{R:NOUN -> NOUN+ADP}). The same UPOS-keyed table fans out to MERLIN for
German and Czech and to UA-GEC for Ukrainian (Table~\ref{tab:jp-merlin-uagec}).
The UPOS substrate intentionally cannot distinguish G/Case, G/Number, and
G/Gender for Ukrainian or G\_Agr from G\_Morphol\_Wrong for German. For each
underdetermined UPOS, the fixed routing shown in
Table~\ref{tab:jp-merlin-uagec} selects one corpus-native target. Pure-UPOS labels make several
MERLIN classes (\texttt{O\_Wordbd}, \texttt{O\_Capit}, \texttt{G\_Agr},
\texttt{G\_Wo}) structurally unreachable, which limits the jp\_errant baseline
independently of the mapping choice.

\begin{table}[h]
\centering
\small
\setlength{\tabcolsep}{4pt}
\begin{tabular}{l l l}
\toprule
\textbf{UPOS} & \textbf{MERLIN (DE/CS)} & \textbf{UA-GEC (UK)} \\
\midrule
NOUN  & G\_Morphol\_Wrong & G/Case \\
PROPN & G\_Morphol\_Wrong & G/Case \\
VERB  & G\_Verb           & G/Tense \\
AUX   & G\_Verb\_compl    & G/Tense \\
ADJ   & G\_Morphol\_Wrong & G/Case \\
ADV   & V\_Wordform       & G/Other \\
ADP   & G\_Prep           & G/Prep \\
DET   & G\_Art            & G/Other \\
PRON  & G\_Refl           & G/Case \\
CCONJ & G\_Conj           & G/Conjunction \\
SCONJ & G\_Conj           & G/Conjunction \\
CONJ  & G\_Conj           & G/Conjunction \\
PART  & G\_Verb\_compl    & G/Particle \\
PUNCT & O\_Punct          & Punctuation \\
NUM   & O\_Abbrev         & G/Other \\
INTJ  & V\_semdenot\_word\_fs & G/Other \\
SYM   & O\_Abbrev         & Punctuation \\
X     & V\_semdenot\_word\_fs & G/Other \\
UNK   & V\_semdenot\_word\_fs & G/Other \\
\bottomrule
\end{tabular}
\caption{jp\_errant\_bea (multilingual ERRANT modification) UPOS category to MERLIN
(used for both DE and CS, which share the MERLIN reference
taxonomy) and to UA-GEC. The pattern body is collapsed to its head UPOS
before lookup.}
\label{tab:jp-merlin-uagec}
\end{table}

For KoLLA, jp\_errant\_bea-ko emits labels in KAGAS form (\texttt{R:SPELL},
\texttt{R:NOUN -> NOUN}, \texttt{R:WO}, etc.), which is the same label space
as the KoLLA reference. No additional mapping is applied. Because
jp\_errant\_bea-ko and the KoLLA reference use the same label space,
near-perfect classification is expected and reflects shared annotation
conventions rather than independent agreement (\S\ref{sec:results-reference}).

\section{spaCy-3.8 rebuild of the German ERRANT modification}
\label{app:boyd-port}

The German ERRANT baseline (gERRANT) used in the reference-matching evaluation
(\S\ref{sec:eval-data}) is a rebuild of \citet{boyd-2018-using}'s original
rule-based annotator. The original targets spaCy 1.x interfaces and could not
run in the spaCy 3.8 environment used here. We retain Boyd's
Damerau--Levenshtein alignment, category rules, STTS mapping, command-line
interface, and $M^2$-format output, while updating model and pipeline access and
replacing TreeTagger with spaCy's German lemmatizer. The released rebuild
records these compatibility changes so the German baseline numbers in
Table~\ref{tab:exp-h} can be reproduced from
public artifacts.

\section{Artifacts, licenses, and dataset statistics}
\label{app:artifacts}

Table~\ref{tab:artifacts} lists the corpora, tools, and models used in this
work with their licenses. Every artifact is used for its intended research
purpose: the corpora for grammatical error correction research, the
error-type systems as evaluation baselines, and the encoder as a base model
for fine-tuning. All evaluation corpora are published research resources. We
did not independently audit their source text for personal or sensitive
content. The released evaluation artifacts include corpus text
and annotations under their original terms.
Human-evaluation records are deidentified and omit
annotator identities and free-form comments.

The METC repository provides the fine-tuned checkpoint,
inference and evaluation code, training configuration,
training-data statistics, annotation and judge prompts,
taxonomy definitions, cross-taxonomy mappings, and
frozen evaluation artifacts. The proprietary training
corpus is not released. The code and model weights
are released under the Apache 2.0 license, while
third-party artifacts retain their respective licenses
and terms of use.

Table~\ref{tab:dataset-stats} reports the size of each evaluation setting.

\begin{table*}[htbp]
\centering
\footnotesize
\begin{tabular}{ll}
\toprule
Artifact & License / terms \\
\midrule
CoNLL-2014 test set & research use (shared task) \\
MERLIN & CC BY-SA 4.0 \\
KoLLA v2 & GPL-3.0-or-later \\
UA-GEC & CC BY 4.0 \\
MuCGEC & Apache 2.0 \\
ERRANT & MIT \\
jp\_errant & research use (authors' terms) \\
diff-match-patch & Apache 2.0 \\
spaCy & MIT \\
Stanza & Apache 2.0 \\
mmBERT-small encoder & MIT \\
METC code       & Apache 2.0 \\
METC checkpoint & Apache 2.0 \\
\bottomrule
\end{tabular}
\caption{Artifacts used in this work and their licenses or terms of use.}
\label{tab:artifacts}
\end{table*}

\begin{table*}[htbp]
\centering
\small
\begin{tabular}{lrr}
\toprule
Setting & Sentences & Reference edits \\
\midrule
EN ann0 & 946 & 2,389 \\
EN ann1 & 1,042 & 3,207 \\
DE & 1,197 & 2,412 \\
CS & 6,590 & 14,143 \\
KO & 1,418 & 1,768 \\
UK & 2,127 & 5,390 \\
ZH & 1,079 & 4,191 \\
\bottomrule
\end{tabular}
\caption{Evaluation dataset statistics: sentences with at least one edit and
reference edit counts per setting.}
\label{tab:dataset-stats}
\end{table*}

The settings also differ in domain and represented population. CoNLL-2014
contains essays by non-native English learners at the National University of
Singapore \citep{ng-etal-2014-conll}. MERLIN contains learner texts from
standardized German and Czech language examinations spanning CEFR levels A1--C1
\citep{boyd-etal-2014-merlin}. KoLLA v2 contains Korean learner writing
with multiple correction references and rubric-based scores
\citep{park-2025-kolla-v2}. UA-GEC includes native and non-native Ukrainian
writing from essays, chats, social media, and formal text
\citep{syvokon-etal-2023-ua}. MuCGEC draws Chinese-as-a-second-language learner
sentences from the PKU learner corpus, proficiency exams, and Lang-8
\citep{zhang-etal-2022-mucgec}. Cross-language differences therefore also
reflect differences in authorship and domain, not language alone.

\section{Use of AI assistants}
\label{app:ai-assistants}

AI assistants (Claude by Anthropic and Codex by OpenAI) were used to draft and
edit prose under author direction, to write analysis and evaluation code, and for LaTeX
mechanics. All analyses, claims, and final wording were reviewed and
validated by the authors.

\clearpage
\begin{table*}[p]
\centering
\small
\begin{tabular}{llrrrrr}
\toprule
Dataset & System & $n$ & Span G & Label G & Label A & Label B \\
\midrule
EN ann0 & Human       &  2,389 & 83.4 & 69.7 & 21.1 &  9.3 \\
EN ann0 & METC       &  2,392 & 74.8 & \textbf{63.7} & 27.4 &  8.9 \\
EN ann0 & METC-teacher & 2,392 & 74.8 & \textbf{64.0} & 26.2 &  9.8 \\
EN ann0 & ERRANT     &  2,447 & 78.6 & 60.7 & 29.1 & 10.2 \\
EN ann0 & jp\_errant &  2,435 & 79.0 & 59.0 & 32.7 &  8.3 \\
\midrule
EN ann1 & Human       &  3,207 & 87.1 & 78.8 & 14.5 &  6.6 \\
EN ann1 & METC       &  3,249 & 77.1 & \textbf{66.7} & 24.2 &  9.1 \\
EN ann1 & METC-teacher & 3,249 & 77.1 & \textbf{66.7} & 25.2 &  8.1 \\
EN ann1 & ERRANT     &  3,343 & 78.9 & 62.5 & 28.7 &  8.9 \\
EN ann1 & jp\_errant &  3,353 & 79.0 & 60.9 & 30.8 &  8.4 \\
\midrule
DE      & Human       &  2,407 & 76.4 & 59.3 & 19.6 & 21.1 \\
DE      & METC       &  2,996 & 79.4 & 72.4 & 18.0 &  9.6 \\
DE      & METC-teacher & 2,996 & 78.8 & 72.1 & 19.8 &  8.1 \\
DE      & gERRANT    &  2,879 & 84.8 & \textbf{75.0} & 20.5 &  4.5 \\
DE      & jp\_errant &  2,814 & 81.8 & 47.8 & 33.0 & 19.2 \\
\midrule
CS      & Human       & 14,125 & 85.2 & 60.5 & 19.0 & 20.5 \\
CS      & METC       & 16,014 & 83.9 & 62.5 & 25.1 & 12.4 \\
CS      & METC-teacher & 16,014 & 84.4 & \textbf{66.4} & 25.4 &  8.2 \\
CS      & jp\_errant & 15,533 & 87.1 & 26.0 & 45.5 & 28.5 \\
\midrule
KO      & KAGAS       &  1,768 & 77.5 & 39.8 & 25.8 & 34.4 \\
KO      & METC       &  1,744 & 70.2 & \textbf{57.3} & 32.3 & 10.4 \\
KO      & METC-teacher & 1,744 & 72.5 & \textbf{58.4} & 32.2 &  9.4 \\
KO      & jp\_errant &  1,768 & 77.6 & 40.1 & 25.5 & 34.4 \\
\midrule
UK      & Human       &  5,167 & 79.4 & 74.3 & 16.8 &  8.9 \\
UK      & METC       &  4,605 & 82.1 & 64.7 & 24.8 & 10.5 \\
UK      & METC-teacher & 4,605 & 81.2 & \textbf{78.7} & 15.0 &  6.3 \\
UK      & jp\_errant &  5,197 & 71.0 & 64.9 & 23.3 & 11.8 \\
\midrule
ZH      & ChERRANT   &  4,191 & 69.1 & 39.9 & 41.6 & 18.5 \\
ZH      & METC       &  4,203 & 67.4 & 48.8 & 32.0 & 19.2 \\
ZH      & METC-teacher & 4,203 & 67.1 & \textbf{50.3} & 34.0 & 15.8 \\
ZH      & jp\_errant &  4,497 & 63.1 & 31.8 & 27.4 & 40.8 \\
\bottomrule
\end{tabular}
\caption{Full primary-judge acceptability results. Values are percentages and are point
estimates. Span G is the share of edits judged Good on span quality. Label G,
A, and B are the shares judged Good, Acceptable, and Bad on label quality
(judge inputs: \S\ref{sec:results-judge}). Bold marks the highest label-Good
rate among the compared systems under the reported judge, together with any
system whose paired sentence-bootstrap difference from it includes zero;
reference rows are not bolded.}
\label{tab:exp-l}
\end{table*}

\begin{table*}[p]
\centering
\small
\begin{tabular}{llrrrr}
\toprule
Dataset & System & Perfect & Acceptable & Bad & Adequacy \\
\midrule
EN ann0 & Human  & 48.7 & 24.8 & 26.4 & 61.2 $\pm$ 2.5 \\
EN ann0 & METC       & 46.3 & 26.0 & 27.7 & \textbf{59.3} $\pm$ 2.6 \\
EN ann0 & METC-teacher & 46.1 & 24.7 & 29.2 & \textbf{58.5} $\pm$ 2.6 \\
EN ann0 & ERRANT     & 42.4 & 26.5 & 31.1 & 55.7 $\pm$ 2.7 \\
EN ann0 & jp\_errant & 40.7 & 28.6 & 30.7 & 55.0 $\pm$ 2.6 \\
\midrule
EN ann1 & Human  & 50.2 & 26.5 & 23.3 & 63.4 $\pm$ 2.4 \\
EN ann1 & METC       & 35.6 & 34.5 & 29.8 & \textbf{52.9} $\pm$ 2.5 \\
EN ann1 & METC-teacher & 36.7 & 32.4 & 30.9 & \textbf{52.9} $\pm$ 2.5 \\
EN ann1 & ERRANT     & 33.1 & 34.0 & 32.9 & 50.1 $\pm$ 2.6 \\
EN ann1 & jp\_errant & 30.1 & 35.4 & 34.5 & 47.8 $\pm$ 2.4 \\
\midrule
DE      & Human  & 40.2 & 14.3 & 45.4 & 47.4 $\pm$ 3.5 \\
DE      & METC       & 50.7 & 20.5 & 28.8 & 60.9 $\pm$ 2.8 \\
DE      & METC-teacher & 49.3 & 23.6 & 27.0 & 61.2 $\pm$ 2.8 \\
DE      & gERRANT    & 52.5 & 26.1 & 21.5 & \textbf{65.5} $\pm$ 2.7 \\
DE      & jp\_errant & 22.4 & 35.1 & 42.6 & 39.9 $\pm$ 2.6 \\
\midrule
CS      & Human  & 35.8 & 16.8 & 47.4 & 44.2 $\pm$ 1.4 \\
CS      & METC       & 37.5 & 29.6 & 32.9 & 52.3 $\pm$ 1.2 \\
CS      & METC-teacher & 40.4 & 33.1 & 26.5 & \textbf{56.9} $\pm$ 1.1 \\
CS      & jp\_errant &  9.9 & 35.6 & 54.5 & 27.7 $\pm$ 0.9 \\
\midrule
KO      & KAGAS  & 23.7 & 23.0 & 53.3 & 35.2 $\pm$ 2.6 \\
KO      & METC       & 33.7 & 32.5 & 33.7 & 50.0 $\pm$ 2.8 \\
KO      & METC-teacher & 37.6 & 33.4 & 29.0 & \textbf{54.3} $\pm$ 2.7 \\
KO      & jp\_errant & 24.0 & 23.1 & 52.9 & 35.5 $\pm$ 2.6 \\
\midrule
UK      & Human  & 41.6 & 30.7 & 27.7 & 57.0 $\pm$ 2.1 \\
UK      & METC       & 32.4 & 36.9 & 30.7 & 50.8 $\pm$ 2.1 \\
UK      & METC-teacher & 40.7 & 37.0 & 22.3 & \textbf{59.2} $\pm$ 2.1 \\
UK      & jp\_errant & 33.4 & 27.8 & 38.9 & 47.3 $\pm$ 2.2 \\
\midrule
ZH      & ChERRANT  & 10.4 & 32.0 & 57.6 & 26.4 $\pm$ 1.9 \\
ZH      & METC       & 18.2 & 22.3 & 59.5 & 29.3 $\pm$ 2.3 \\
ZH      & METC-teacher & 19.1 & 25.9 & 55.1 & \textbf{32.0} $\pm$ 2.3 \\
ZH      & jp\_errant &  7.9 & 15.0 & 77.1 & 15.4 $\pm$ 1.8 \\
\bottomrule
\end{tabular}
\caption{Full primary-judge sentence-level outcomes and adequacy scores aggregated from the
per-edit judgments in Table~\ref{tab:exp-l}. Perfect, Acceptable, and Bad are
percentages of judged sentences; adequacy has a 95\% sentence-bootstrap
interval shown as $\pm$ its half-width. Reference is included as a baseline,
not a competing system. Bold
marks the highest-scoring compared system per setting under the reported
judge, together with any system whose paired sentence-bootstrap difference
from it includes zero.}
\label{tab:exp-sent}
\end{table*}

\end{document}